%% file: main.tex
\documentclass{article}
\usepackage[final]{colm2026_conference}

\usepackage{microtype}
\usepackage{hyperref}
\usepackage{url}
\usepackage{booktabs}

\usepackage[T1]{fontenc}    
\usepackage{url}            
\usepackage{amsfonts}       
\usepackage{nicefrac}       
\usepackage{microtype}      
\usepackage{amssymb,amsthm}
\usepackage{amsmath}
\usepackage{mathtools}
\usepackage{makecell}
\usepackage{pifont}
\usepackage{multirow}
\usepackage{comment}
\usepackage{xspace}
\usepackage{float}
\usepackage{wrapfig}
\usepackage{enumitem}
\usepackage{todonotes}

\usepackage{lineno}

\definecolor{darkblue}{rgb}{0, 0, 0.5}
\hypersetup{colorlinks=true, citecolor=darkblue, linkcolor=darkblue, urlcolor=darkblue}

\title{Domain-Aware Scaling Laws Uncover Data Synergy}

\author{
 Kimia Hamidieh$^1$\thanks{
 Correspondence to \texttt{hamidieh@mit.edu}.}
    \quad Lester Mackey$^2$
    \quad David Alvarez-Melis$^{2, 3}$ \\ \\
    \normalsize $^1$MIT CSAIL, $^2$ Microsoft Research, $^3$ Harvard University
}

\begin{document}

\ifcolmsubmission
\linenumbers
\fi

\maketitle

\raggedbottom

    \begin{abstract}
        Machine learning progress is often attributed to scaling model size and dataset volume, yet the composition of data can be just as consequential. Empirical findings repeatedly show that combining datasets from different domains yields nontrivial interactions. For instance, adding code improves mathematical reasoning, while certain mixtures introduce interference that reduces model performance. We refer to these effects collectively as data \textit{synergy}, where the contribution of multiple domains exceeds or falls short of the sum of their isolated contributions. In this work, we formalize and quantify data synergy in language model pretraining. Leveraging observational variation across open-weight LLMs with diverse pretraining mixtures, we estimate both direct domain-to-benchmark synergy (how one domain contributes to performance on another) and a second-order domain-domain synergy (capabilities that require co-occurrence of multiple domains). Our framework improves predictive accuracy over domain-agnostic scaling laws and recovers stable synergy estimates. We validate these estimates by training models on predicted optimal and predicted anti-optimal mixtures and confirm that our synergy estimates correctly predict performance rankings.\looseness-1
    \end{abstract}

    \input{sections/introduction}

    \input{sections/scaling_law}
    \input{sections/second_order}

    \input{sections/experimental_setup}
    \input{sections/results}

    \input{sections/related_works}
    \input{sections/conclusion.tex}

\newpage
{
\setlength{\bibsep}{6pt}
\bibliography{bibl}
\bibliographystyle{colm2026_conference}
}

\newpage
\appendix
\input{sections/appendix}

\end{document}

%% file: sections/introduction.tex
\section{Introduction}

Recent improvements in Large Language Models (LLMs) are strongly shaped by their pretraining data distribution~\citep{li2024datacomp,soldaini2024dolma}, yet prior work abstracts away composition and reduces data to an undifferentiated token count~\citep{kaplan2020scaling,hoffmann2022training}. In practice, however, pretraining corpora are mixtures of data from different domains (e.g., web, books, code, math) and small shifts in composition can significantly impact model capabilities~\citep{ye2024data,liu2024regmix}. In particular, empirical studies repeatedly report interactions between data domains that are systematic rather than anecdotal: code pretraining can improve mathematical~\citep{lu2024mathcoder2,azerbayev2023llemma} and logical~\citep{yu2024codepmp} reasoning, math and code mixtures often outperform either alone~\citep{aryabumi2024code}, while other combinations lead to interference and degrade performance on certain tasks~\citep{zheng2024beyond,li2024improving,gu2024cmr}. These observations suggest that tokens are not interchangeable. What matters is not only how much data we train on, but also \textit{what kinds of data are combined}. 

\begin{figure}[t]
    \centering
    \vspace{-.3in}
    \includegraphics[width=\linewidth]{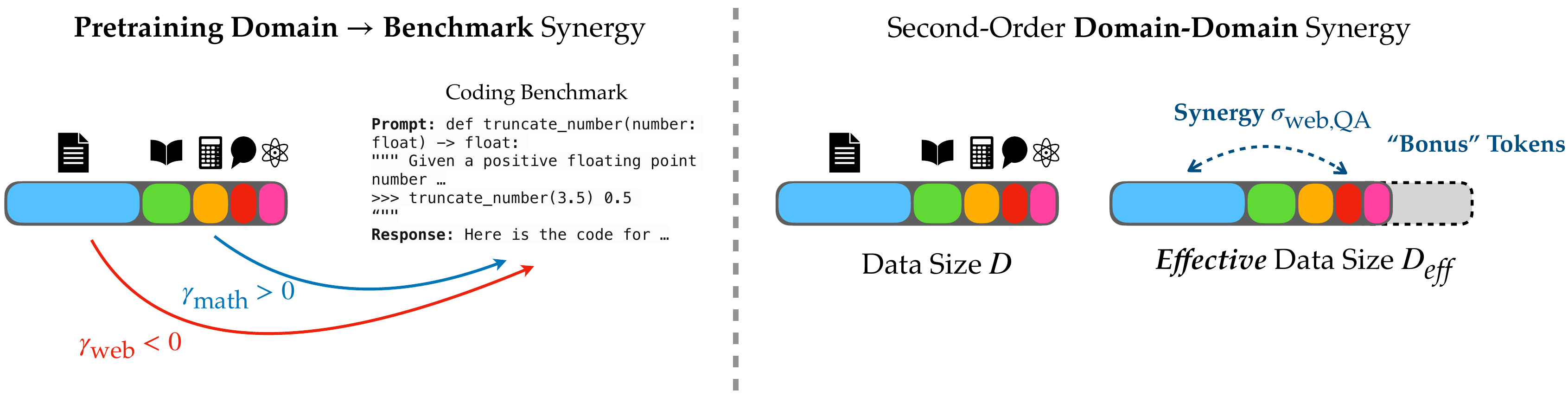}
    \vspace{-.3in}
    \caption{\textbf{Two forms of data synergy:} Pretraining corpora are mixtures of different data domains (Web, Books, Math, Q\&A, Science, \dots), and their \emph{composition} shapes downstream performance.
    \emph{\textbf{Left:}} Each domain modifies the \emph{rate} at which additional data reduces loss on a given benchmark, which is captured by a per-domain adjustment $\gamma_\text{domain}$ to the data scaling exponent. On a coding benchmark, adding math data reduces loss faster, while web data reduces it more slowly.
    \emph{\textbf{Right:}} When two synergistic pretraining domains co-occur, their pairwise interaction $\sigma_{kk'}$ contributes additional ``bonus'' tokens that vanish if either domain is absent, so the corpus behaves as if it had a larger \emph{effective} data size $D_\textit{eff} > D$. These synergies are shared across benchmarks, but their bonus tokens stay benchmark-specific.
    }
    \label{fig:two_forms_synergy}
    \vspace{-.1in}
\end{figure}

We refer to these interactions collectively as \textbf{data synergy}. We distinguish two complementary forms (Figure~\ref{fig:two_forms_synergy}). The first is \emph{domain$\to$benchmark synergy}: the extent to which pretraining data from one domain helps or hurts performance on a given benchmark, beyond what would be expected from total token count. The second is \emph{domain-domain synergy}: non-additive effects that arise when two domains co-occur in the training mixture, so that their joint contribution is greater or smaller than the sum of their isolated effects. The former captures the effect of a training domain on a given benchmark or evaluation domain, and the latter captures higher-order complementarities or interference internal to the pretraining corpus itself.

Most existing approaches overlook data synergy: either they treat pretraining corpora as homogeneous, or they model composition without modeling the interactions between domains. Classical scaling laws, for instance, relate loss to parameter count and total data but are domain-agnostic, since they assume all tokens contribute equally~\citep{kaplan2020scaling,hoffmann2022training}. Similarly, data attribution and curation methods usually learn per-example or per-domain contributions to a target metric~\citep{bae2024training,grosse2023studying,ilyas2022datamodels}, and mixture optimization methods often search over weights under an implicit assumption of independent domains~\citep{xie2023doremi,liu2024regmix}. What is missing is a formalization of data synergy that separates the aggregate benefit of more data from the effect of its composition, and quantifies when two domains together yield more (or less) than the sum of their parts.

In this paper, we present \textbf{domain-aware scaling laws}, a framework that quantifies and models data synergy in LLM pretraining. 
Rather than training a large controlled set of models, we leverage observational variation across open-weight LLMs with diverse pretraining mixtures. 
Our framework modifies standard scaling laws with additional terms that capture both first-order domain$\to$benchmark synergy and second-order synergy between pretraining domains. 
The resulting estimators improve predictive accuracy over domain-agnostic scaling laws and recover sparse, interpretable estimates of synergy and interference. For example, we find recurring positive interactions between code and math, alongside negative interactions for some domain pairs that consistently degrade performance across different benchmarks.
Our contributions are as follows:

\begin{itemize}[topsep=0pt, parsep=0pt, itemsep=0pt, partopsep=0pt]
    \item We provide a formal, quantifiable definition of data synergy that 
    captures prior empirical observations.
    \item We introduce domain-aware scaling laws with first- and second-order synergy coefficients that improve predictive model performance over standard scaling laws.
    \item We recover stable, interpretable synergy estimates that provide actionable insights for data curation and acquisition.
    \item We validate the framework by training models on predicted-optimal and predicted-anti-optimal mixtures and show that observationally derived synergy estimates correctly predict performance rankings across multiple benchmarks.
\end{itemize}

%% file: sections/scaling_law.tex
\section{Domain and Synergy-Aware Scaling Laws}
\label{sec:domain_scaling}

In this section, we develop scaling laws that progressively incorporate training data composition and synergy into the standard scaling laws. We begin with the problem setup and a domain-agnostic baseline in Sections~\ref{sec:problem_setup} and~\ref{sec:domain_agnostic}, then extend the data term to capture first-order domain$\to$benchmark synergy in Section~\ref{sec:first_order_synergy} and second-order synergy between pretraining domains in Section~\ref{sec:cooccurance_synergy}.

\subsection{Problem Setup}
\label{sec:problem_setup}

Let $ \{M_1, \dots, M_m\} $ denote a set of language models (e.g., open-weight LLMs on Huggingface), and let $ \{ D_1, \dots, D_n \}$ be evaluation domains. For every model-domain pair we observe the loss value
$L=\bigl[l_{i,j}\bigr] \in \mathbb{R}^{m\times n}$
where rows $L_{i,:}$ summarize the performance of model $M_i$ across domains and columns $L_{:,j}$ compare models on domain $D_j$.
Alongside $L$ we collect statistics of model $M_i$, such as parameter count $N_i$, total pretraining tokens $d_i$, and composition (mixture shares) of the pretraining data. Let $u_{i,k} \in [0,1]$ be the fraction of tokens from training domain $D_k$ that model $M_i$ is pretrained on, with $\sum_k u_{i,k}=1$ and $d_i=\sum_k u_{i,k}d_i$.

\begin{figure}[t]
    \centering
    \vspace{-.35in}
    \includegraphics[width=\linewidth]{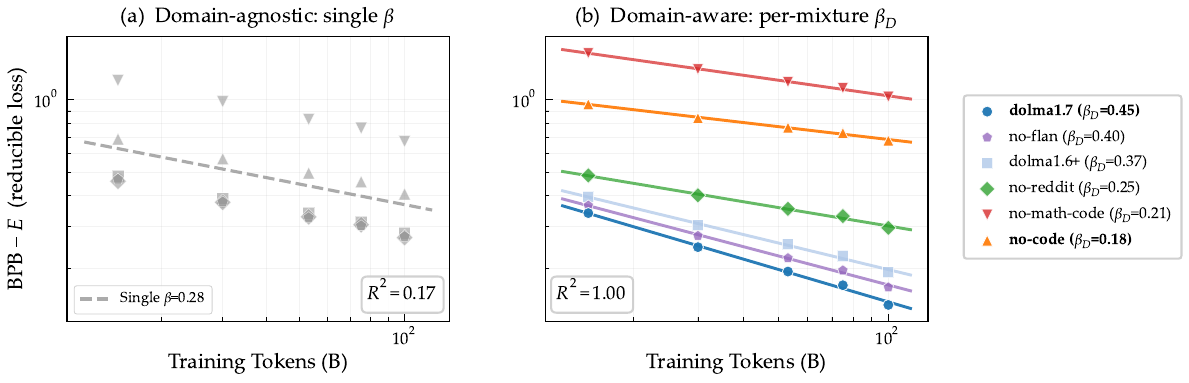} 
    \vspace{-0.25in}
    \caption{\textbf{Data scaling exponent depends on pretraining mixture}: a motivating example on \texttt{HumanEval}. \emph{\textbf{(a)}} A single domain-agnostic scaling exponent ($\beta=0.28$) fits the data poorly ($R^2=0.17$), as models with different mixtures show different reducible losses at a fixed token count. \emph{\textbf{(b)}} Allowing a per-mixture exponent $\beta_D$ captures the variation from different data mixtures ($R^2=1.00$). Removing code or math data \emph{reduces} the data scaling exponent (the slope here), while the full mixture scales faster.
    }
    \vspace{-0.1in}
    \label{fig:motivation_fanout}
\end{figure}

Our goal is to quantify \emph{data synergy across domains}, or how training on pretraining domain(s) affects loss on a benchmark, after accounting for model scale and total tokens.

\subsection{Domain-agnostic scaling law}
\label{sec:domain_agnostic}

We begin with a domain-agnostic scaling law that explains loss variation using only model size $N$ in terms of number of parameters and total pretraining tokens $D$. This is the baseline model against which composition- and synergy-aware adjustments will be evaluated. 
Following Chinchilla scaling laws~\citep{hoffmann2022training}, we have the parametric form
\begin{equation}
    L(N,D)=L_\infty + A N^{-\alpha}+B D^{-\beta},
    \label{eq:chinchilla}
\end{equation}
where $L_\infty>0$ is the irreducible loss, $A,B>0$ are scale coefficients for parameter and data terms, and $\alpha,\beta>0$ are the parameter and data scaling exponents. Typical values of the exponents are $\alpha \approx 0.06$ and $\beta \approx 0.20$.

In practice we fit this baseline \emph{per benchmark} $j$ and allow $(L_{\infty,j},A_j,B_j,\alpha_j,\beta_j)$ to vary across evaluation domains, after transforming benchmark performance metrics to a common pseudo-loss scale.
Let $s=\log N$ and $d=\log D$, and define $e_j=\log L_{\infty,j}$, $a_j=\log A_j$, and $b_j=\log B_j$. Writing \eqref{eq:chinchilla} in log parameters gives the numerically-stable log-sum-exp (LSE) form:
\begin{equation}
    \log L_j(s,d)=\operatorname{LSE}\!\big(e_j,\; a_j-\alpha_j s,\; b_j-\beta_j d\big),
    \label{eq:domain_agnostic}
\end{equation}
where $\operatorname{LSE}(x_1,x_2,x_3)=\log \bigl(e^{x_1}+e^{x_2}+e^{x_3}\bigr)$.
Equation~\eqref{eq:domain_agnostic} is our \emph{domain-agnostic} baseline: the expected log-loss depends only on total parameters and total tokens.
In the next subsections we expand and modify the \emph{data} term $b - \beta\log D$ to account for the data mixture and to estimate synergies, while preserving the overall form of the scaling law.

\subsection{First-order Domain-Benchmark Synergy}
\label{sec:first_order_synergy}

We now modify the data term so that \emph{different training domains can reduce loss at different rates} on each benchmark, as illustrated in Figure~\ref{fig:motivation_fanout}.
Recall that $u_{i,k} \in [0,1]$ is the fraction of tokens from training domain $D_k$ that model $M_i$ is pretrained on. 
It is straightforward to show that:
\begin{equation}
    \log d_i=\sum_k u_{i,k}\log(u_{i,k}d_i)+H(u_i),
    \label{eq:log_d_decomp}
\end{equation}
where $H(u_i)=-\sum_k u_{i,k}\log u_{i,k}$. 
We introduce domain-specific data scaling exponent parameters $\{\gamma_{j,k}\}$, which modify the data scaling exponents per-task additively: $\beta_j + \gamma_{j,k}$. A positive $\gamma_{j,k}$ means that tokens from domain $k$ reduce loss on benchmark $j$ faster than the baseline rate $\beta_j$ predicts, which we refer to as a \emph{synergistic} effect. Conversely, a negative $\gamma_{j,k}$ implies \emph{interference}: tokens from domain $k$ reduce loss more slowly than expected. 
Substituting~\eqref{eq:log_d_decomp} into the third argument of the LSE in
\eqref{eq:domain_agnostic} and introducing these modifiers we have:
\begin{align}
    \mathbb{E}[\,l_{i,j}\,]
    =\operatorname{LSE} \Bigl(
        e_j, 
        a_j-\alpha_j s_i, 
        b_j-\bigl[\beta_j\mathbf 1+\gamma_{j,\cdot}\bigr]^{ \top}z_i-\beta_j H(u_i)
    \Bigr),
    \label{eq:domain_benchmark_lse}
\end{align}
with $z_{i,k} = u_{i,k} \log(u_{i,k}d_i)$ and synergy coefficients $\gamma_{j,k}\in\mathbb{R}$. Equivalently, in the original (non-log) parameterization, this is a multiplier to the data term in \eqref{eq:chinchilla},
\begin{equation}
    L_j(N_i, d_i, u_i) = L_{\infty,j} + A_j N_i^{-\alpha_j} + B_j\, d_i^{-\beta_j}\prod_k (u_{i,k} d_i)^{-\gamma_{j,k} u_{i,k}},
    \label{eq:domain_benchmark_product}
\end{equation}
so each domain $k$ rescales the effective data term by a factor $(u_{i,k} d_i)^{-\gamma_{j,k} u_{i,k}}$ that depends on both its token share $u_{i,k}$ and its absolute token count $u_{i,k} d_i$.

%% file: sections/second_order.tex
\subsection{Second-order Pretraining Data Synergy}
\label{sec:cooccurance_synergy}

We hypothesize that certain training domains interact synergistically, such that their joint presence yields learning benefits independent of the specific downstream benchmark. These synergies reflect fundamental complementarities between domains, though their effect size is still mediated by the data scaling coefficient of each benchmark. The gain from such co-occurrence is bottlenecked by the scarcer domain and can be understood as producing ``additional bonus tokens'' only when both domains are present. To capture this, we model synergy with a pairwise term that vanishes if either domain is absent and scales with the smaller per-domain log-token budget.

Define
\[
z_{i,k}=u_{i,k}\log(u_{i,k}d_i),\quad
\bar z_{i,k}=\log \bigl(1+u_{i,k}d_i\bigr),\quad
\operatorname{softmin}_\tau(a,b):=-\tau\log \bigl(e^{-a/\tau}+e^{-b/\tau}\bigr).
\]

Start from the first-order data term split into baseline and per-domain parts,
$
-\beta_j\sum_{k} z_{i,k}\;-\;\sum_{k}\gamma_{j,k}\,z_{i,k}\;-\;\beta_j H(u_i),
$
and modify only the first part with a co-occurrence term
\[
-\beta_j\sum_{k} \Bigl[z_{i,k}
    +\sum_{k'\ne k}\gamma_{j,k}\sigma_{kk'}\,\operatorname{softmin}_\tau\bigl(\bar z_{i,k},\bar z_{i,k'}\bigr)\Bigr]\;-\;\sum_{k}\gamma_{j,k}z_{i,k}
    \;-\;\beta_j H(u_i),
\]
which is a sum over pairwise synergies $\sigma_{kk'}\,\operatorname{softmin}_\tau\bigl(\bar z_{i,k},\bar z_{i,k'}\bigr)$ modulated by the domain-benchmark synergy strength.

Symmetrizing (using $\sigma_{kk'}=\sigma_{k'k}$) gives
\[
-\sum_{k}(\beta_j+\gamma_{j,k})\,z_{i,k}
\;-\;
\beta_j\sum_{k<k'}(\gamma_{j,k}+\gamma_{j,k'})\,\sigma_{kk'}\,
\operatorname{softmin}_\tau\bigl(\bar z_{i,k},\bar z_{i,k'}\bigr)
\;-\;\beta_j H(u_i).
\]
Using $\bar z$ (nonnegative and $=0$ when $u_{i,k}=0$) ensures that the interaction truly vanishes if either domain is absent and preserves the desired $O(\log d_i)$ scaling. The pairwise term carries the same $\beta_j$ as the baseline, so it moves at the same rate and reads as a change in \emph{effective data count}.
Without the $\beta_j$ factor, $\sigma_{kk'}$ would be harder to interpret across benchmarks with different $\beta_j$.

\paragraph{Formulation.}
Starting from the LSE formulation of the baseline scaling law, plugging in the new data term we have
\begin{align}
\Phi_{i,j}
&= b_j
   -\bigl[\beta_j\mathbf 1_K+\gamma_{j,\cdot}\bigr]^{ \top}z_i
   -\beta_j\,H(u_i)
   -\beta_j \sum_{k<k'} (\gamma_{j,k}+\gamma_{j,k'})\,\sigma_{kk'}\;
      \operatorname{softmin}_\tau \bigl(\bar z_{i,k},\bar z_{i,k'}\bigr),
\label{eq:second_order_phi}
\end{align}
where $\Sigma=[\sigma_{kk'}]$ is symmetric with $\sigma_{kk}=0$. The complete log-space law has the form
\[
\log L_{i,j}
=\operatorname{LSE} \bigl(e_j,\;a_j-\alpha_js_i,\;\Phi_{i,j}\bigr).
\]

Intuitively, the pairwise term can be understood as producing benchmark-specific \emph{effective tokens}. Define $\log D^{\mathrm{eff}}_{i,j} := \log d_i + \sum_{k<k'}(\gamma_{j,k}+\gamma_{j,k'})\,\sigma_{kk'}\,s_{i,kk'}$, where $s_{i,kk'} = \operatorname{softmin}_\tau(\bar z_{i,k},\bar z_{i,k'})$. Reordering the terms to achieve a view similar to Chinchilla, the data term scales with $(D^{\mathrm{eff}}_{i,j})^{-\beta_j}$. This shows that the co-occurrence of synergistic domains contributes ``bonus tokens'' that increase the effective dataset size for a given benchmark, while interfering pairs reduce it. A complete derivation is provided in Appendix~\ref{app:sec_syn}.

%% file: sections/experimental_setup.tex
\begin{figure}[t]
    \centering
    \vspace{-.3in}
    \includegraphics[width=\linewidth]{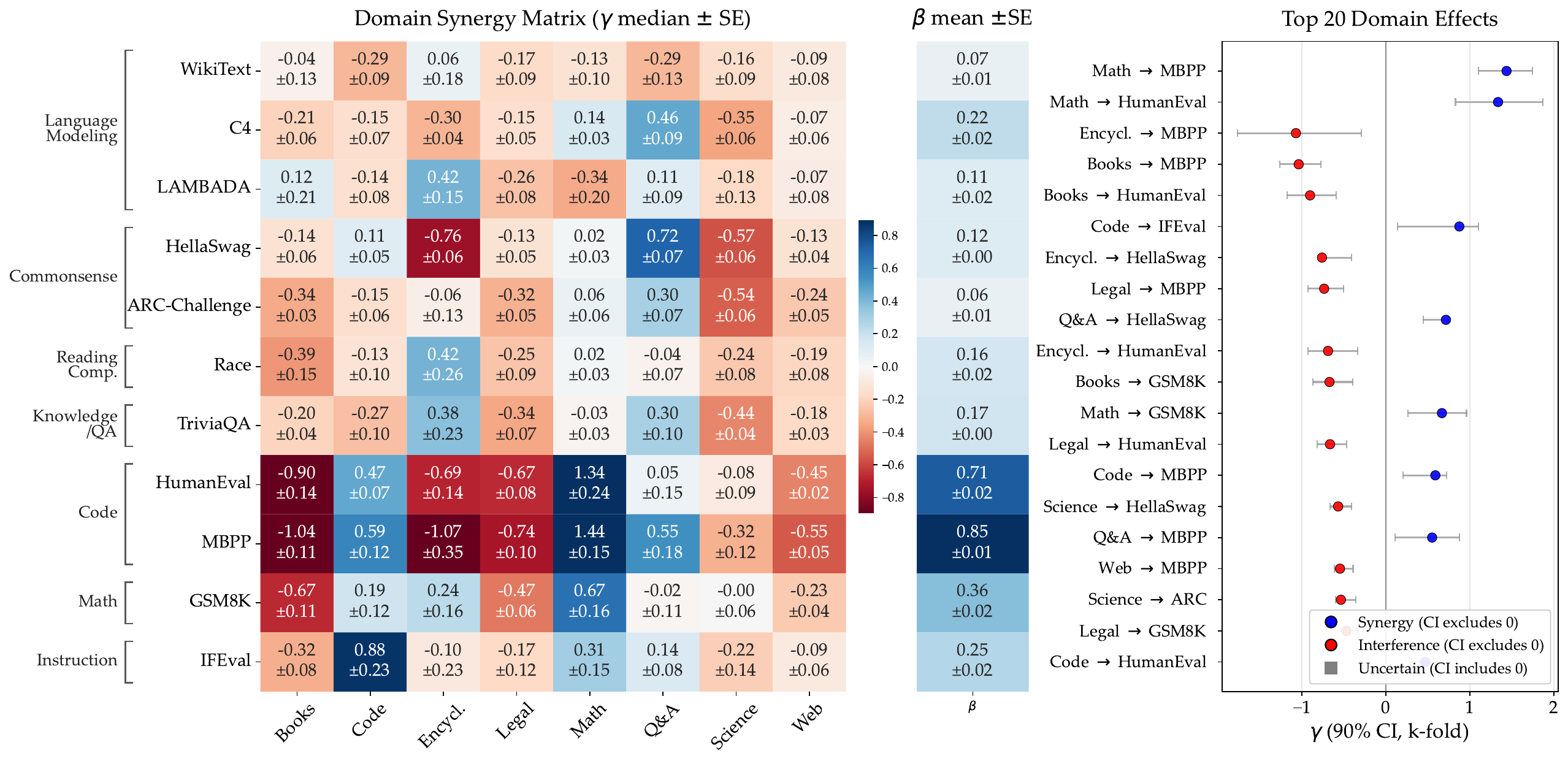}
    \vspace{-.3in}
    \caption{\textbf{Overview of estimated domain$\to$benchmark synergy.} We fit domain-aware scaling laws across open-weight LLMs and extract per-domain synergy coefficients ${\gamma}_{j,k}$ that quantify how each pretraining domain contributes to benchmark performance beyond aggregate token count. \emph{\textbf{Left:}} Heatmap of ${\gamma}_{j,k}$ (median $\pm$ SE) between pretraining domains (columns) and benchmarks (rows), where positive values show synergy and negative values visualize interference. \emph{\textbf{Center:}} Benchmark-specific data exponent ${\beta}_j$. \emph{\textbf{Right:}} Top 20 first-order synergies ranked by magnitude, with 90\% bootstrap confidence intervals.}
    \label{fig:estimated_gamma}
    \vspace{-.1in}
\end{figure}

\section{Estimation from Observational Data}
\label{sec:setup}

A distinguishing feature of our approach is that all scaling law parameters, including domain synergies, are estimated from \emph{observational} data, a collection of publicly available pretrained language models with documented pretraining mixtures, rather than models trained in controlled experiments. Because the data is observational rather than interventional, the synergy coefficients we estimate should be understood as associations, i.e., they capture how data composition varies with benchmark performance across the observed model set, but do not by themselves establish causal domain interactions. We take a step toward causal validation in Section~\ref{sec:pretrain}.
This broad observational coverage comes with several estimation challenges, which we address through normalization, regularization, cross-validation, and uncertainty quantification. We summarize the data, evaluation choices, and fitting methodology below, and details are deferred to Appendices~\ref{app:training_uncertainty},~\ref{app:obs_data}, and~\ref{app:fitting_obs}.

\textbf{Data.}
Our model set spans six open-weight model groups (\texttt{GPT-Neo/J/NeoX}~\citep{gpt-neo,gpt-j,black2022gpt}, \texttt{Pythia}~\citep{biderman2023pythia}, \texttt{DataDecide}~\citep{magnusson2025datadecide}, \texttt{OLMo}~\citep{groeneveld2024olmo}, \texttt{OpenLLaMA}~\citep{openlm2023openllama}, \texttt{RedPajama-INCITE}~\citep{weber2024redpajama}), 52 models in total from 70M to 20B parameters. We map each model's pretraining sources to eight domains (\emph{books}, \emph{code}, \emph{encyclopedia}, \emph{legal}, \emph{math}, \emph{Q\&A}, \emph{science}, \emph{web}). These domains are aligned with the Data Provenance Initiative~\citep{longpre2023data} and Essential-Web v1.0~\citep{essentialweb2025} data categories. The 30 \texttt{DataDecide} models add controlled pretraining data ablations (\emph{no-code}, \emph{no-math-code}, \emph{no-flan}, \emph{no-reddit}) that are especially informative for estimating synergies. We evaluate on 11 benchmarks: \texttt{HumanEval}~\citep{chen2021evaluating}, \texttt{MBPP}~\citep{austin2021program}, \texttt{GSM8K}~\citep{cobbe2021training}, \texttt{ARC-Challenge}~\citep{clark2018think}, \texttt{HellaSwag}~\citep{zellers2019hellaswag}, \texttt{RACE}~\citep{lai2017race}, \texttt{TriviaQA}~\citep{joshi2017triviaqa}, \texttt{LAMBADA}~\citep{paperno2016lambada}, \texttt{IFEval}~\citep{zhou2023instruction}, \texttt{WikiText}~\citep{merity2016pointer}, and \texttt{C4}~\citep{raffel2019exploring}.
Because many models are small, exact match scores on open-ended tasks are often near zero, so we score those tasks by bits-per-byte (BPB) of the gold response and rank-Gaussian transform~\citep{conover1981rank} all targets to a common scale before fitting.

\textbf{Identifiability.}
In observational data, $\log D$ and $\log N$ are correlated, so the Chinchilla exponents $\alpha$ and $\beta$ trade off freely. However, the composition features $z_{i,k} = u_{i,k}\log(u_{i,k}d_i)$ are nearly orthogonal to $\log N_i$, which means the synergy coefficients $\gamma$ are well-identified even when $\alpha$ and $\beta$ individually are not. An important caveat is that the observed mixture vectors are effectively low-dimensional, so we interpret $\gamma_{j,k}$ as composition sensitivities along the observed mixture axes rather than as fully causal synergy estimates (Appendix~\ref{app:fitting_obs}). 

\textbf{Robust fitting.}
We fit both the first-order (FO) and second-order (SO) models in two stages. First, we fit a per-task Chinchilla baseline~\eqref{eq:chinchilla}, then the full synergy model. We place a prior on each $\beta_j$ centered at its Chinchilla value $\hat\beta_j^{\text{chin}}$, which keeps the data scaling exponent positive and lets the synergy terms capture composition, so the fit stays interpretable at no cost in accuracy. Fitting the non-convex LSE objective on a small number of models is prone to overfitting, so we use a Huber loss~\citep{huber1992robust} for robustness and $\ell_1$ and $\ell_2$ regularization to make the $\gamma_{j,k}$ sparse under collinear domain fractions. We minimize the loss using the L-BFGS algorithm~\citep{nocedal1980updating}.

\textbf{Uncertainty and model selection.}
We evaluate the fit by 5-fold cross-validation over the models and quantify uncertainty by refitting on $B{=}50$ bootstrap resamples (80\% of models). We consider only synergy entries whose 90\% intervals exclude zero as meaningful. The FO model is fit per task, while the SO model is fit jointly since $\Sigma$ is shared across tasks. We select fits by \emph{parameter stability}, the consistency of $\gamma$ across folds, alongside held-out $R^2$ (Appendix~\ref{app:training_uncertainty}).

%% file: sections/results.tex
\section{Results}
We present our findings in three parts. First, we estimate the first-order domain$\to$benchmark synergies (Section~\ref{sec:first_order_results}). Second, we estimate the second-order pairwise interactions between pretraining domains (Section~\ref{sec:second_order_results}). Finally, we validate these results through controlled mixture-optimization experiments (Section~\ref{sec:pretrain}).

\begin{figure}[t]
    \centering
    \vspace{-.3in}
    \includegraphics[width=\linewidth]{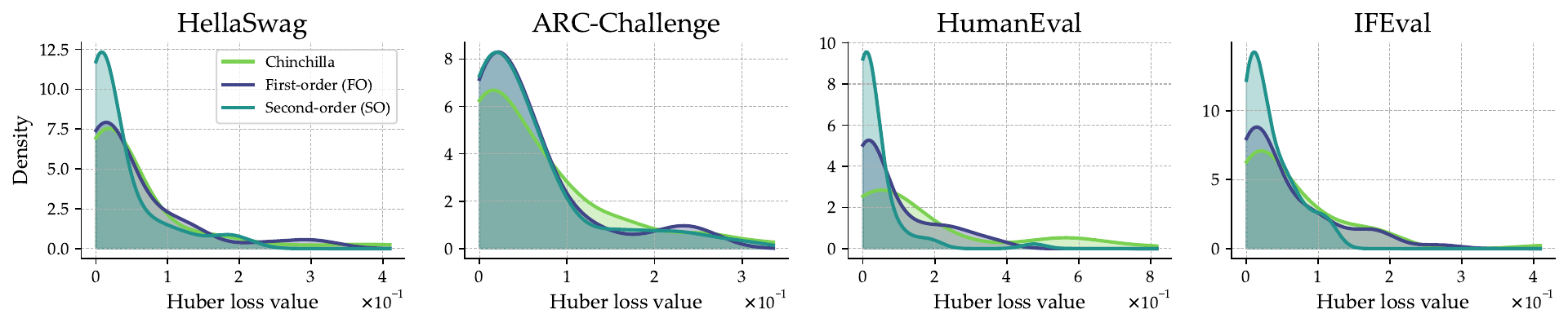}
    \vspace{-.15in}
    \caption{\textbf{Distribution of held-out prediction loss for three fitted scaling models}: We evaluate the Huber loss between each fitted scaling law's prediction and the observed value on held-out test examples, for the domain-agnostic Chinchilla baseline and the first-order (FO) and second-order (SO) synergy models, across four benchmarks. Both synergy-aware models concentrate more density near zero, which means that they achieve more accurate predictions on unseen models. The largest improvements appear on \texttt{HumanEval} and \texttt{IFEval}.}
    \vspace{-.15in}
    \label{fig:syn_chinchilla_comparison}
\end{figure}

\subsection{First-order Domain$\to$Benchmark Synergy}
\label{sec:first_order_results}

We first estimate the first-order synergy of each pretraining domain on each benchmark, which captures how each domain contributes to benchmark performance beyond aggregate token count.
Our first-order (FO) domain-aware model adds a domain coefficient $\gamma_{j,k}$ to each benchmark's data scaling exponent, $\beta_j + \gamma_{j,k}$ \eqref{eq:domain_benchmark_lse}: a positive $\gamma_{j,k}$ means tokens from domain $k$ increase the scaling rate on benchmark $j$ above its baseline $\beta_j$. We fit this model to the observational data in Section~\ref{sec:setup}. 

\paragraph{Strongest synergies.}
Figure~\ref{fig:estimated_gamma} shows the estimated matrix $\Gamma=\{\gamma_{j,k}\}$, the data scaling exponents $\beta_j$, and the top 20 synergies ranked by magnitude, each with bootstrap standard errors and 90\% confidence intervals. The strongest synergies appear for code and math benchmarks. For example, \texttt{HumanEval} shows $\gamma_{\text{code}} = {+}0.47$ and $\gamma_{\text{math}} = {+}1.34$, and \texttt{MBPP} shows $\gamma_{\text{code}} = {+}0.59$ and $\gamma_{\text{math}} = {+}1.44$. These values are consistent with prior findings that math and code data reinforce each other on programming benchmarks~\citep{azerbayev2023llemma,lu2024mathcoder2}. 
The synergies are robust across cross-validation folds and sign-consistent for most benchmark-domain pairs (Appendix~\ref{app:training_uncertainty}), and the same fit also identifies reliable interference, most clearly from books and encyclopedia on code benchmarks. Composition affects not only the benchmark performance but the rate at which it scales with data from each domain. Figure~\ref{fig:motivation_fanout} shows the data exponent $\beta$ on \texttt{HumanEval} falls from $0.45$ under the full \texttt{dolma1.7} mixture to $0.18$ once code is removed (Appendix~\ref{app:per_family_scaling}).

\paragraph{First-order fit.}
The FO model obtains a median cross-validated $R^2$ of $0.906$ across benchmarks. Its gain over the domain-agnostic Chinchilla baseline is largest on the code benchmarks, where the total token count alone explains little of the variance: the baseline reaches only $R^2 = 0.41$ on \texttt{HumanEval} and $0.20$ on \texttt{MBPP}, while the FO model reaches $0.92$ and $0.88$ (Appendix Figure~\ref{fig:perf_comparison}).

\subsection{Second-order Pretraining Data Synergy}
\label{sec:second_order_results}
Beyond first-order effects, we estimate \emph{second-order} data synergy between pretraining domains, which captures gains that materialize only when two domains co-occur. The SO model introduces shared pairwise domain interaction parameters $\sigma_{kk'}$, estimated jointly across all benchmarks.
We fit the symmetric synergy matrix $\Sigma=\{\sigma_{kk'}\}$ along with other parameters as in \eqref{eq:second_order_phi} and evaluate uncertainty by bootstrap resampling.

\begin{figure}[t]
    \centering
    \vspace{-0.3in}
    \includegraphics[width=1.0\linewidth]{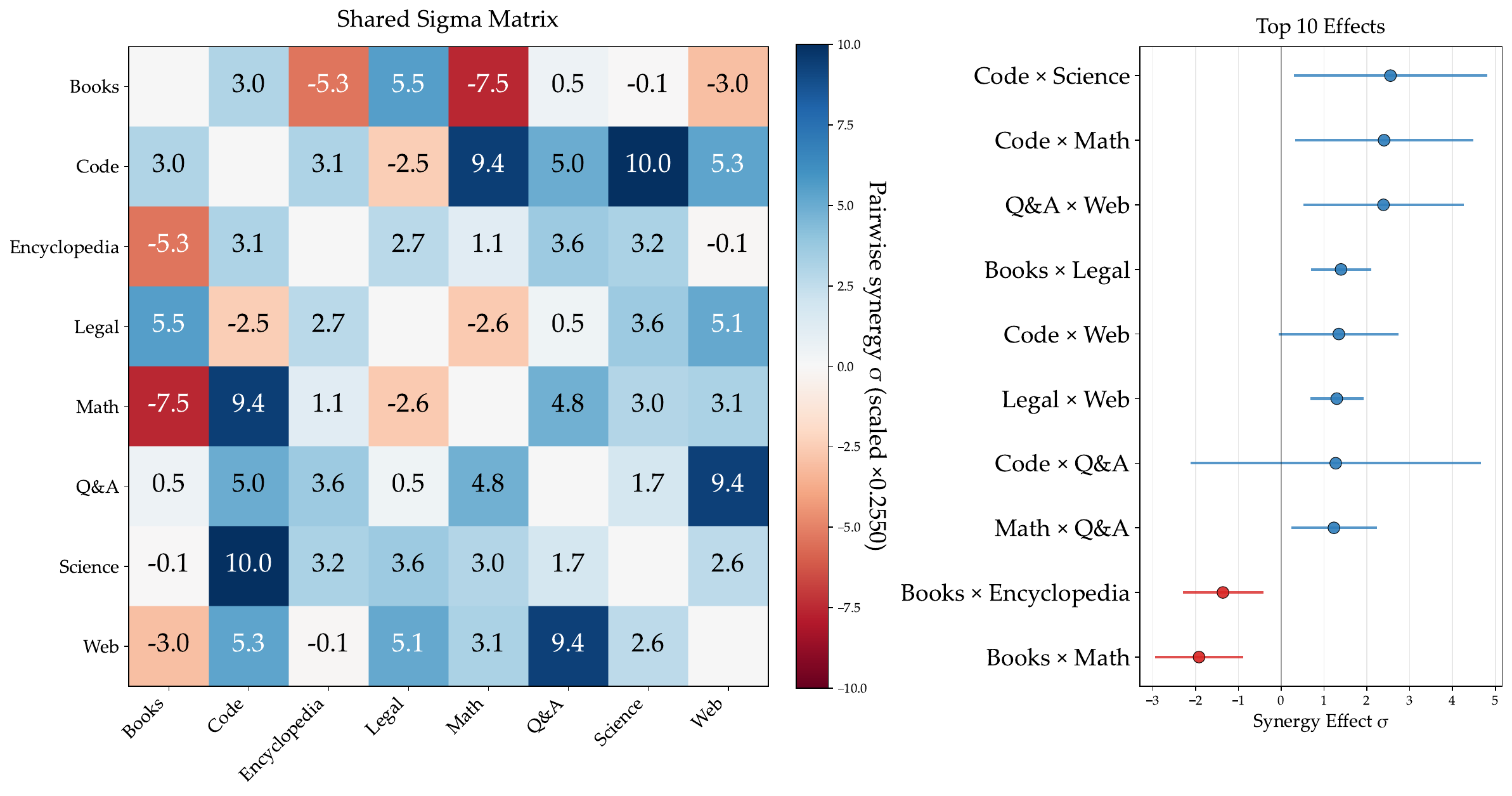}
    \vspace{-0.35in}
    \caption{
    \textbf{Second-order domain-domain synergy. }
    \textbf{\emph{Left:}} The shared pairwise interaction matrix $\Sigma = \{\sigma_{kk'}\}$, where positive entries show synergy between two domains and negative entries show interference. \textbf{\emph{Right:}} The largest second-order synergies with 90\% bootstrap confidence intervals. Code shows the strongest positive synergy with math and science data, which appears only when the domains appear together in pretraining data.
    }
    \label{fig:estimated_sigma}
    \vspace{-.1in}
\end{figure}

\paragraph{Strongest pairwise synergies.} Figure~\ref{fig:estimated_sigma} summarizes the shared pairwise synergy matrix $\Sigma=\{\sigma_{kk'}\}$ across domains, along with the confidence intervals. The largest pairwise interactions are $\sigma_{\text{code,science}} = {+}2.55$ and $\sigma_{\text{code,math}} = {+}2.40$. When code co-occurs with math or scientific data in the pretraining mixture, the pair contributes more than the sum of the two domains' first-order effects. 
The clearest negative interaction is \emph{Math}$\times$\emph{Books}: their co-presence in the pretraining corpus may lower average performance on the affected benchmarks. This suggests that the two domains share little structure, so tokens spent on both dilute each other rather than provide the mutual reinforcement seen for code and math. 
A negative $\sigma$ is therefore actionable, when a target benchmark favors one such domain, a mixture should not also spend tokens on its interfering domains.

\paragraph{Second-order fit.} The SO model obtains a median cross-validated $R^2$ of $0.912$ across benchmarks and improves on FO most where FO already showed strong domain effects: \texttt{GSM8K} ($+0.013$), \texttt{IFEval} ($+0.020$), and \texttt{TriviaQA} ($+0.019$). Its per-benchmark $\gamma$ coefficients shrink relative to FO (Appendix Figure~\ref{fig:second_order_gamma}), because the shared $\sigma$ absorbs part of the domain$\to$benchmark synergy.
On held-out prediction loss (Figure~\ref{fig:syn_chinchilla_comparison}), both the FO and SO models place more of the loss distribution near zero than the domain-agnostic baseline, so the fitted $\gamma_{j,k}$ and $\sigma_{kk'}$ generalize rather than overfit, most clearly on benchmarks unlike typical pretraining text such as \texttt{HumanEval} (Appendix~\ref{app:comparison} reports all fits). Additionally, our first- and second-order laws also achieve the best held-out fit among prior composition-aware scaling laws (Appendix~\ref{app:prior_comparison}).

\subsection{Validation via Mixture Optimization}
\label{sec:pretrain}

The synergy estimates above are observational. To test whether they are \emph{actionable}, we use the domain-aware scaling laws along with estimated first-order synergies to predict benchmark performance, and leverage those predictions for pretraining mixture optimization. We train models at two scales, 30M and 150M parameters, on mixtures predicted to be optimal or anti-optimal for three target tasks: \texttt{HumanEval}, \texttt{GSM8K}, and \texttt{IFEval}. Appendix~\ref{app:optim_math} shows that the predicted-optimal mixture puts the most weight on the domains with the largest positive $\gamma_{j,k}$ (e.g., \emph{math} and \emph{code} for \texttt{HumanEval}) and the least on the rest, and the anti-optimal mixture does the reverse (e.g., \emph{books} and \emph{encyclopedia}). We optimize over the domains from \texttt{DataDecide}, and the balanced baseline gives equal weight to each domain. Full details of this experiment are in Appendix~\ref{app:validation}.

\begin{table}[h]
\centering
\vspace{-.15in}
\caption{Bits-per-byte (BPB, $\downarrow$) on target tasks at two model scales. The scaling law correctly predicts the ranking between optimal and anti-optimal for all tasks at both scales. Percentages are relative to the balanced baseline.}
\label{tab:bpb_results}
\small
\begin{tabular}{lcccc}
\toprule
Metric (BPB) & Balanced & Optimal & Anti-optimal & Gap \\
\midrule
\multicolumn{5}{l}{\textit{30M parameters, 3B tokens}} \\
\texttt{HumanEval} & 1.268 & \textbf{1.110} ($-$12.4\%) & 1.503 (+18.6\%) & 0.393 \\
\texttt{GSM8K}     & 1.818 & \textbf{1.779} ($-$2.1\%)  & 1.935 (+6.5\%)  & 0.156 \\
\texttt{IFEval}    & \textbf{1.451} & 1.513 (+4.3\%)    & 1.551 (+6.9\%)  & 0.038 \\
\midrule
\multicolumn{5}{l}{\textit{150M parameters, 5B tokens}} \\
\texttt{HumanEval}
& 0.868 & \textbf{0.721} ($-$16.9\%) & 1.058 (+21.9\%) & 0.337 \\
\texttt{GSM8K}     & 1.643 & \textbf{1.402} ($-$14.7\%) & 1.591 ($-$3.2\%)  & 0.189 \\
\texttt{IFEval}    & 1.366 & \textbf{1.324} ($-$3.0\%)  & 1.365 ($-$0.0\%)  & 0.041 \\
\bottomrule
\end{tabular}
\end{table}

Table~\ref{tab:bpb_results} reports bits-per-byte on each task's held-out set. At 30M, the predicted-optimal mixture outperforms the anti-optimal on all three tasks, which confirms the scaling law's directional prediction. The gap is largest on \texttt{HumanEval}, where the optimal mixture obtains a 31\% BPB improvement (relative to the balanced baseline) over the anti-optimal.

At 150M, the predictions still hold for all three tasks, and the relative gaps widen: \texttt{HumanEval} keeps a large gap (0.337 BPB), \texttt{GSM8K} strengthens from a 2.1\% to a 14.7\% reduction over balanced, and \texttt{IFEval} stays small but consistent. The synergy coefficients $\gamma_{j,k}$, estimated purely from observational data, thus carry actionable insight for mixture curation that persists as model scale increases.

%% file: sections/related_works.tex
\section{Related Work}

\textbf{Scaling laws.} LLM pretraining validation loss follows power laws as model capacity and data grow. Early results characterize scaling with parameters, data, and compute \citep{kaplan2020scaling}, while later work refines the compute-data trade-off and recommends that token counts scale roughly with parameters to remain compute-optimal~\citep{hoffmann2022training}. Recent work fits such laws \emph{observationally}, from existing public models: a low-dimensional capability space predicts downstream performance~\citep{ruan2024observational}, and fitted parameters transfer across models that differ in architecture and data~\citep{choshen2024hitchhiker}. 
We treat these domain-agnostic laws as our baseline and add terms that depend on data composition to scaling laws.

\textbf{Domain-aware scaling and mixture optimization.}
Beyond total token count, several works show \emph{which} tokens matter. Data pruning can reduce error to exponential decay in dataset size given a good pruning metric~\citep{sorscher2022beyond}. Mixture optimization methods select the domain mixture that minimizes loss through regression~\citep{liu2024regmix} or parametric mixing laws that capture nonlinear returns~\citep{ye2024data}, and online methods further adapt the mixture during training through unified reweighting~\citep{chen2024aioli} or gradient-based balancing~\citep{ge2025r}. 
Closest to our work, a few methods fit variations of scaling laws to select the mixture~\citep{kang2024autoscale,shukor2025scaling,longpre2025atlas}, but they keep the data scaling exponent global or use additive per-domain power laws instead of modifying the data term. Our first-order model instead embeds composition into the scaling exponent itself ($\beta_j{+}\gamma_{j,k}$, which varies by both domain and benchmark), so that mixture changes modify the \emph{rate} at which additional data reduces the loss. Our second-order extension further introduces explicit pairwise domain interactions ($\sigma_{kk'}$) that none of these methods capture.

\textbf{Observational inference of skills and performance.}
A complementary literature utilizes observational variation across many (model, task) points: perplexity correlations guide data selection without training~\citep{thrush2024improving}, latent variable models recover shared capability factors~\citep{jin2025discovering}, and data influence methods show that group interactions can cancel or amplify individual effects~\citep{yu2025group}. We share this spirit but predict performance under counterfactual mixtures from domain-level composition.
Other related work predicts downstream performance from pretraining loss or compute~\citep{chen2024scaling,schaeffer2025pretraining}, including on generative benchmarks.

\textbf{Evidence for data synergy.}
Multiple empirical studies report positive transfer between code and mathematical reasoning. Continued pretraining that combines math and code improves math benchmarks~\citep{azerbayev2023llemma,lu2024mathcoder2,wang2026mergemix}, code during pretraining rather than only at the SFT stage improves broader reasoning with little negative transfer~\citep{ma2023training}, and code pretraining improves entity tracking~\citep{kim2024code}. These findings motivate sparse, domain-specific parameters that capture synergy between conceptually related domains.

%% file: sections/conclusion.tex
\section{Conclusion \& Discussion}
\label{sec:conclusion}
We formalize and quantify data synergy in LLM pretraining and show that explicitly modeling domain-benchmark and second-order synergy between training domains recovers interpretable, sparse synergy estimates and improves predictive fit across multiple benchmarks.
We also note that one could discover synergies by changing the definition of domains, e.g., different modalities, different languages, or finer splits of code and science. We estimate synergy in pretraining, but whether the same domain pairs stay synergistic under supervised fine-tuning or reinforcement learning remains open. 

Our approach is limited by its observational design, as estimates rely on potentially noisy or incomplete mixture metadata, and the effective mixture space is low-dimensional. The validation experiments remain at small scale, so verification at larger model sizes is an important direction for future work. Beyond methodology, our synergy estimates support practical applications in data curation, mixture design, and data acquisition, and offer a principled tool for targeting specific capabilities in future training runs.

%% file: sections/appendix.tex
\section{Second-order Synergy as Effective Tokens}

\label{app:sec_syn}

For a given benchmark $j$ and a model $i$ trained with a given data composition, define the \emph{effective tokens}
\[
\log D^{\mathrm{eff}}_{i,j}
:=\log d_i\;+\;\sum_{k<k'}(\gamma_{j,k}+\gamma_{j,k'})\,\sigma_{kk'}\,s_{i,kk'},
\quad
D^{\mathrm{eff}}_{i,j}=d_i\exp \Bigl(\sum_{k<k'}(\gamma_{j,k}+\gamma_{j,k'})\,\sigma_{kk'}\,s_{i,kk'}\Bigr),
\]
so that, by \eqref{eq:log_d_decomp},
\[
\exp(\Phi_{i,j})
=\exp \bigl(b_j-\gamma_{j,\cdot}^{ \top}z_i\bigr)\;\bigl(D^{\mathrm{eff}}_{i,j}\bigr)^{-\beta_j}.
\]
Written in the same format as the Chinchilla scaling law~\eqref{eq:chinchilla}, we have:
\[
L_{i,j}(N_i,d_i)\;\ = \;
L_{\infty,j}+A_j N_i^{-\alpha_j}
+\underbrace{\widetilde{B}_{i,j}}_{=\exp(b_j-\gamma_{j,\cdot}^{ \top}z_i)}\;
\bigl(D^{\mathrm{eff}}_{i,j}\bigr)^{-\beta_j}.
\]
For small interactions, the first-order approximation $e^{-x}\approx 1-x$ gives
\[
\bigl(D^{\mathrm{eff}}_{i,j}\bigr)^{-\beta_j}
\ \approx
d_i^{-\beta_j} \left[
1-\beta_j \sum_{k<k'}(\gamma_{j,k}+\gamma_{j,k'})\,\sigma_{kk'}\,\operatorname{softmin}_\tau(\bar z_{i,k},\bar z_{i,k'})\right],
\]
which explicitly shows how the data term changes with ``bonus tokens'' from the co-occurrence of two domains, modulated by the universal $\Sigma$. For a pair of domains with positive $\sigma_{kk'}$ and $\gamma_{j,k}+\gamma_{j,k'}$, the data term in the loss decreases, and we recover the first-order form if all entries in $\Sigma$ are zero.

\section{Training and Uncertainty}
\label{app:training_uncertainty}

This appendix provides full details on the fitting procedure, hyperparameter choices, and uncertainty quantification summarized in Section~\ref{sec:setup}.

\paragraph{Two-stage fitting procedure.}
We fit both first-order (FO) and second-order (SO) synergy scaling laws in two stages. The FO model is fit independently for each of the 11 tasks, while the SO model is fit jointly across tasks because the pairwise matrix $\sigma_{kk'}$ is shared across benchmarks. All evaluation targets are rank transformed before fitting, for robustness to outliers and to different loss scales across tasks.

\emph{Stage 1 (Chinchilla baseline).}
For each benchmark $j$, we fit the domain-agnostic Chinchilla model \eqref{eq:chinchilla} independently, which gives per-task estimates $(\hat\alpha_j^{\text{chin}}, \hat\beta_j^{\text{chin}}, \hat e_j, \hat a_j, \hat b_j)$. The fitted $\hat\beta_j^{\text{chin}}$ then center the Stage~2 $\beta$ prior.

\emph{Stage 2 (synergy model).}
We fit the full synergy model, first-order~\eqref{eq:domain_benchmark_lse} or second-order~\eqref{eq:second_order_phi}, for each task.

\paragraph{Objective.}
We minimize a Huber$_\delta$ risk~\citep{huber1992robust} over the LSE loss (Section~\ref{sec:domain_scaling}):
\begin{align}
\min_{\Theta}\ &\sum_{j=1}^{n}\sum_{i=1}^{m}
\operatorname{Huber}_{\delta}\Bigl(
\operatorname{LSE}\bigl(e_j,\ a_j-\alpha_j s_i,\ \Phi_{i,j}\bigr)-l_{i,j}\Bigr) \notag \\
&+\;
\lambda_{1}\sum_{j}\|\gamma_{j,\cdot}\|_{1}
\;+\;
\lambda_{2}\sum_{j}\|\gamma_{j,\cdot}\|_{2}^{2}
\;+\;
\lambda_{\beta}\sum_{j}(\beta_j - \hat{\beta}_j^{\text{chin}})^2,
\label{eq:full_objective}
\end{align}
where $s_i=\log N_i$ and $\Phi_{i,j}$ is either from the FO data term \eqref{eq:domain_benchmark_lse} (with $\Sigma=0$) or the SO data term \eqref{eq:second_order_phi}. The Huber loss ($\delta=0.5$) provides robustness to outliers from noisy mixture metadata. The $\ell_1$ regularizer encourages sparse synergy parameters, reflecting the prior that most domain$\to$benchmark interactions are negligible, while $\ell_2$ shrinkage stabilizes estimation under domain collinearity. The SO model shares the pairwise interaction matrix $\sigma_{kk'}$ across tasks (parameterized on the upper triangle only, with $\sigma_{kk}=0$). The reported reference runs use no explicit $\sigma$ penalty.

\paragraph{$\beta$ prior.}
The term $\lambda_{\beta}(\beta_j - \hat{\beta}_j^{\text{chin}})^2$ places a prior on the data scaling exponent, centered on the Chinchilla value from Stage~1. Without it, $\gamma$ and $\beta$ can trade off: the optimizer may reach an equally good fit where $\beta < 0$ and $\gamma$ absorbs the data scaling signal. This prior keeps $\beta_j$ a positive data scaling exponent, so that $\gamma$ captures only domain deviations and both stay interpretable. It acts as a soft constraint during optimization and is the only term that couples the two stages.

\paragraph{Optimization and initialization.}
The FO and SO synergy fits use the same grid for initialization. We grid search over the 5 structural parameters $(e, a, b, \alpha, \beta)$ on a uniform grid,
\begin{align*}
e &\in \{-1.5,\, -1.0,\, -0.5,\, 0.0,\, 0.5,\, 1.0\}, \\
a &\in \{0,\, 5,\, 10\},\quad b \in \{0,\, 5,\, 10\}, \\
\alpha &\in \{0,\, 0.25,\, 0.5,\, 0.75,\, 1.0,\, 1.25\},\quad \beta \in \{0,\, 0.5,\, 1.0\},
\end{align*}
which gives 972 candidates. We subsample 200 at random with a fixed seed and select the initialization with lowest Huber loss. For the FO reference fits, each sampled grid point is optimized with L-BFGS~\citep{nocedal1980updating} using Strong Wolfe line search, history size 30, and up to 5000 iterations. For the SO reference fit, the grid is used to warm-start each task with $\sigma=0$ using the same line search and history size for up to 1000 iterations per grid point, followed by a joint L-BFGS fit over all task parameters and the shared $\sigma_{kk'}$ matrix with history size 50 and up to 5000 iterations. We do \emph{not} initialize Stage~2 from the Chinchilla parameters, as the grid search explores the loss landscape independently. The $\beta$ prior above is the reason we need Stage~1.

\paragraph{Hyperparameters.}
Table~\ref{tab:hyperparams} summarizes the shared and model-specific hyperparameters for the reported reference runs.

\begin{table}[H]
\centering
\small
\setlength{\tabcolsep}{6pt}
\caption{Hyperparameter settings for the first-order (FO) and second-order (SO) reference fits. Rows in the top block are shared and span both columns. Rows in the bottom block give the FO and SO values. A dash means the setting does not apply.}
\label{tab:hyperparams}
\begin{tabular}{lcc}
\toprule
Hyperparameter & FO & SO \\
\midrule
Huber threshold $\delta$ & \multicolumn{2}{c}{0.5} \\
$\ell_2$ regularizer $\lambda_2$ (on $\gamma$) & \multicolumn{2}{c}{0.05} \\
$\beta$ prior $\lambda_\beta$ & \multicolumn{2}{c}{1.0} \\
Grid candidates tried & \multicolumn{2}{c}{200 / 972} \\
Grid line search & \multicolumn{2}{c}{Strong Wolfe} \\
Bootstrap subsample fraction & \multicolumn{2}{c}{80\%} \\
Bootstrap replicates $B$ & \multicolumn{2}{c}{50} \\
\midrule
$\ell_1$ regularizer $\lambda_1$ (on $\gamma$) & $3 \times 10^{-3}$ & $10^{-3}$ \\
Grid L-BFGS history / max iter. & 30 / 5000 & 30 / 1000 \\
Joint L-BFGS history / max iter. & --- & 50 / 5000 \\
Softmin temperature $\tau$ & --- & 0.001 \\
$\sigma$ regularization & --- & 0 \\
\bottomrule
\end{tabular}
\end{table}

\paragraph{Task selection.}
Not all benchmarks exhibit scaling-law behavior. Tasks where the Chinchilla baseline achieves negative $R^2$ (no monotonic relationship between $(N, D)$ and performance) are excluded, as they would contaminate the shared $\sigma$ matrix in SO.

\paragraph{Parameter stability.}
To select among fitting configurations, we use \emph{parameter stability} (PS) alongside held-out $R^2$. PS is defined as the ratio
\[
\text{PS} = \frac{\|\bar{\gamma}\|^2}{\|\bar{\gamma}\|^2 + \|\sigma_\gamma\|^2},
\]
where $\bar\gamma$ is the mean $\gamma$ vector across CV folds and $\sigma_\gamma$ is the standard deviation across folds. A high PS means that $\gamma$ is consistent across data splits, not just that the model predicts well on held-out data. PS and the penalized lower bound on $R^2$ are nearly uncorrelated ($r = 0.048$), which confirms that they capture complementary aspects of fitted model quality: parameter reliability and predictive accuracy, respectively. In our reported first-order fit, PS $= 88\%$, so the estimated $\gamma$ vector is stable across cross-validation folds.

\paragraph{Uncertainty via bootstrap.}
We use the same bootstrap template for FO and SO. First, we fit the model on the full data. For each replicate, we subsample 80\% of model rows without replacement, draw a bootstrap sample with replacement from that subsample, initialize from the full-data solution, and refit. In SO, the same sampled model indices are used for all tasks in a replicate so that the shared $\sigma$ matrix is estimated from aligned rows. We then initialize a final all-data refit from the bootstrap medians and report percentile intervals for $(e_j,a_j,b_j,\alpha_j,\beta_j,\gamma_{j,\cdot})$ and, for SO, the shared coefficients $\sigma_{kk'}$. Only entries whose interval excludes zero are reported as reliable synergy signals in the main text.

\section{Observational Data: Models and Data}
\label{app:obs_data}
We estimate synergies from \emph{observational variation} across open-weight and open-data models and their publicly available pretraining mixtures.
This provides variation in both scale and composition and lets us fit the domain-aware scaling laws, which require \emph{model performance}, parameter and data size, and \emph{data mixture} for each model. 

\begin{figure}[h]
    \centering
    \includegraphics[width=\linewidth]{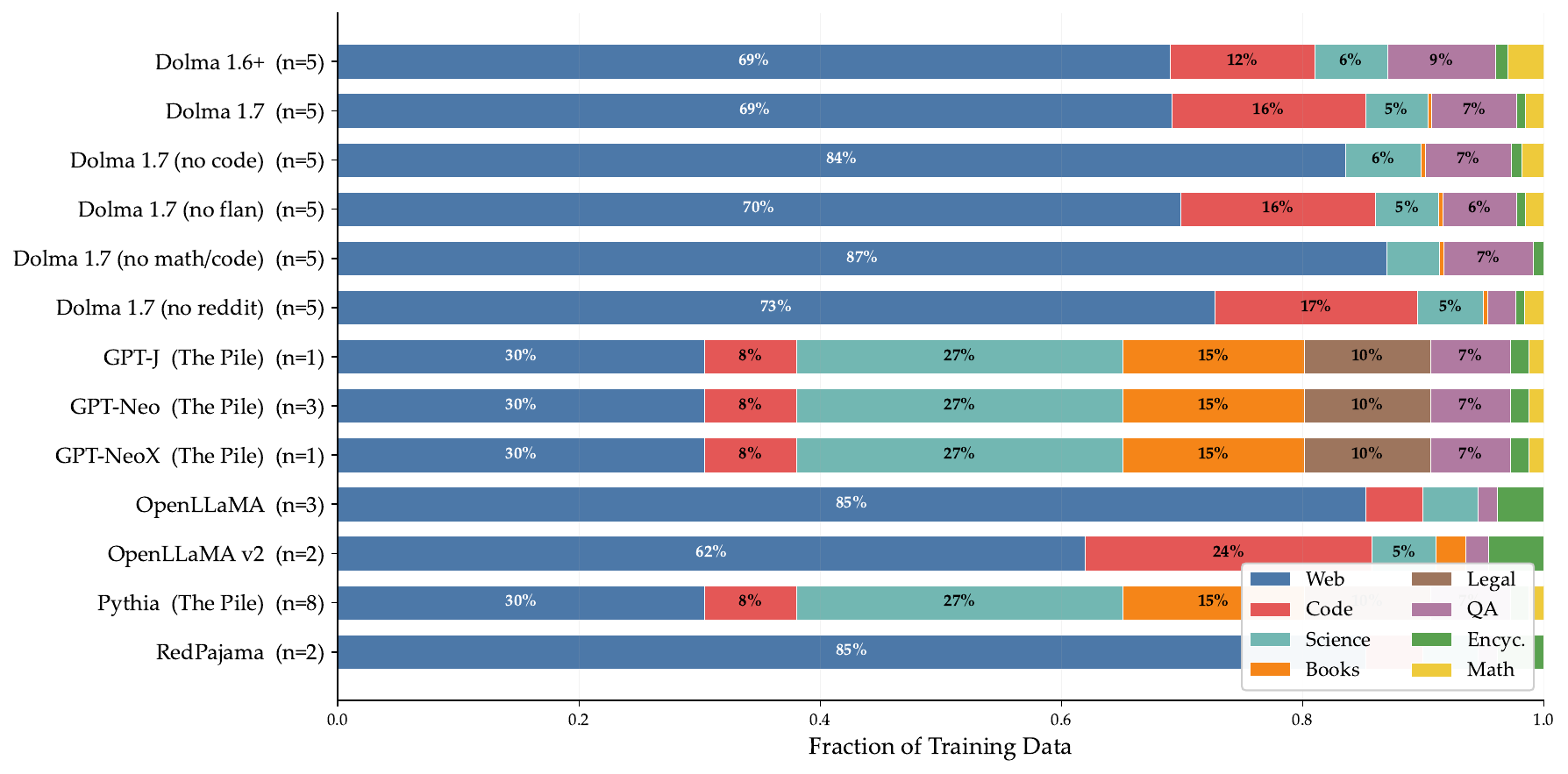}
    \vspace{-0.25in}
    \caption{\textbf{Domain composition of pretraining data for each model group.} The groups trained on the Pile (\texttt{Pythia}, \texttt{GPT-Neo/J/NeoX}) share a single diverse mixture, while the Dolma and other groups are mostly \emph{web}. \texttt{DataDecide} ablations (\texttt{no-code}, \texttt{no-math/code}, \texttt{no-flan}, \texttt{no-reddit}) provide controlled differences in data composition.}
    \label{fig:family_fractions}
\end{figure}

\paragraph{Model groups.}
The main text (Section~\ref{sec:setup}) summarizes the model set. Here we give the per-group scales and checkpoint counts. Our model set spans six open-weight model groups with different scales and training mixtures: \texttt{GPT-Neo/J/NeoX}~\citep{gpt-neo,gpt-j,black2022gpt} (125M--20B, 5 checkpoints), \texttt{Pythia}~\citep{biderman2023pythia} (70M--12B, 8 checkpoints), \texttt{DataDecide}~\citep{magnusson2025datadecide} (150M--1B, 30 checkpoints across Dolma variants with ablations on data composition: \emph{no-code}, \emph{no-flan}, \emph{no-math-code}, \emph{no-reddit}), \texttt{OLMo}~\citep{groeneveld2024olmo} (1B--7B, 2 checkpoints), \texttt{OpenLLaMA}~\citep{openlm2023openllama} (3B--13B, 5 checkpoints), and \texttt{RedPajama-INCITE}~\citep{weber2024redpajama} (3B--7B, 2 checkpoints), totaling 52 checkpoints. The DataDecide ablations vary composition at fixed scale, which helps separate domain effects.
Figure~\ref{fig:family_fractions} shows the resulting data mixture for each model group.

\paragraph{Domain taxonomy.}
Our eight domains align with two independent pretraining taxonomies: the source domains of the Data Provenance Initiative (DPI)~\citep{longpre2023data} and the Level~1 categories of Essential-Web v1.0~\citep{essentialweb2025}. Six domains (\emph{Books, Code, Encyclopedia, Legal, Science, Web}) match DPI source domains. We add \emph{Math} for the explicit math datasets (DM Mathematics, OpenWebMath, Algebraic Stack) that match Essential-Web's math category, and \emph{Q\&A} for sources that are exclusively questions and answers in the mixture metadata, such as StackExchange. The framework is agnostic to the partition: $\gamma_{j,k}$ and $\sigma_{kk'}$ can be re-fit on any categorization, including finer Essential-Web Level~2 or Level~3 categories, languages, or data modalities, provided the mixture metadata identifies those categories reliably.

\paragraph{Dataset to domain mapping.}
For each model group we map published source labels deterministically to the 8 domains (Table~\ref{tab:domain_mapping}). The mapping covers 100\% of tokens, with no ``other'' bucket.

\begin{table}[H]
\centering
\small
\caption{Representative source-to-domain assignments.}
\label{tab:domain_mapping}
\begin{tabular}{ll}
\toprule
Domain & Pretraining sources \\
\midrule
\emph{Books}        & Books3, Project Gutenberg, BookCorpus2 \\
\emph{Code}         & GitHub (Pile), The Stack, StarCoder \\
\emph{Encyclopedia} & Wikipedia, Wikibooks \\
\emph{Legal}        & FreeLaw, USPTO Backgrounds \\
\emph{Math}         & DM Mathematics, OpenWebMath, Algebraic Stack \\
\emph{Q\&A}         & StackExchange \\
\emph{Science}      & ArXiv, PubMed Central, PubMed Abstracts, PhilPapers, NIH ExPorter, peS2o \\
\emph{Web}          & Pile-CC, OpenWebText2, CommonCrawl, C4, RefinedWeb \\
\bottomrule
\end{tabular}
\end{table}

\paragraph{Benchmarks and metrics.}
We evaluate on eleven benchmarks that span a range of capabilities: code generation with \texttt{HumanEval}~\citep{chen2021evaluating} and \texttt{MBPP}~\citep{austin2021program}, mathematical reasoning with \texttt{GSM8K}~\citep{cobbe2021training}, science reasoning with \texttt{ARC-Challenge}~\citep{clark2018think}, commonsense inference with \texttt{HellaSwag}~\citep{zellers2019hellaswag}, reading comprehension with \texttt{RACE}~\citep{lai2017race}, open-domain QA with \texttt{TriviaQA}~\citep{joshi2017triviaqa}, cloze completion with \texttt{LAMBADA}~\citep{paperno2016lambada}, instruction following with \texttt{IFEval}~\citep{zhou2023instruction}, and language modeling with \texttt{WikiText}~\citep{merity2016pointer} and \texttt{C4}~\citep{raffel2019exploring}. All evaluations use the lm-evaluation-harness~\citep{eval-harness}.

Because our model set includes many small models (70M--1B), metrics such as pass@1 and accuracy are close to zero for some generative tasks. For example, 55\% of the models score exactly zero on \texttt{HumanEval} pass@1, and 52\% on \texttt{MBPP}. With the majority of observations at zero, there is no signal for a power law fit. We therefore evaluate generative tasks (\texttt{HumanEval}, \texttt{MBPP}, \texttt{GSM8K}, \texttt{TriviaQA}, \texttt{IFEval}) via bits-per-byte (BPB) of the gold response conditioned on the prompt, computed via log-likelihood. BPB is continuous and monotonically improves with scale: the Chinchilla baseline achieves $R^2 = {-}0.11$ on \texttt{HumanEval} with pass@1, but $R^2 = 0.41$ with BPB. For the remaining six benchmarks (\texttt{ARC-Challenge}, \texttt{HellaSwag}, \texttt{RACE}, \texttt{LAMBADA}, \texttt{WikiText}, \texttt{C4}), accuracy or perplexity already provides a reasonable metric to for fitting scaling laws.

\paragraph{Loss as a proxy for task score.}
For the five generative tasks we fit BPB, so it is worth asking whether lower loss means higher score. Across the models, BPB and the task metric (pass@1 or accuracy) are negatively correlated on all five, but not equally (Table~\ref{tab:bpb_acc_corr}). The correlation is weakest on code and math, where many small models score zero on pass@1 while their BPB keeps improving, so the score stays flat where the loss does not. This is the metric discontinuity that motivates a continuous target~\citep{schaeffer2023emergent,du2024understanding}. Lower loss does not guarantee a higher score, so we fit the metric directly where it is reliable and use BPB only where it is not.

\begin{table}[H]
\centering
\small
\caption{Spearman correlation between BPB and the task metric (pass@1 or accuracy) for the five generative tasks, which is largely negative.}
\vspace{.1in}
\label{tab:bpb_acc_corr}
\begin{tabular}{lccccc}
\toprule
 & \texttt{MBPP} & \texttt{HumanEval} & \texttt{GSM8K} & \texttt{TriviaQA} & \texttt{IFEval} \\
\midrule
Spearman $\rho$ & $-0.39$ & $-0.54$ & $-0.62$ & $-0.69$ & $-0.72$ \\
\bottomrule
\end{tabular}
\end{table}

\paragraph{Metric normalization.}
To place different metrics such as error rate and BPB on a common scale, we apply a rank-Gaussian (inverse normal) transform $g_j(m)=\Phi^{-1}\!\bigl(\hat F_j(m)\bigr)$, where $\hat F_j$ is the empirical CDF of metric values and $\Phi^{-1}$ is the standard normal quantile function, which gives pseudo log-losses $l_{i,j}=g_j(m_{i,j})\propto\log L_{i,j}$.

\section{Robustness and Validity of Synergy Estimates}
\label{app:fitting_obs}

Because we fit from observational data rather than a set of models trained with controlled sizes and data mixtures, we test whether the estimated synergies are robust or artifacts of the models we have selected. We first describe the structure of the observed mixtures, then check the synergies against four factors that could otherwise explain them: the low dimensional mixture space, data repetition, model scale, and overlap between domains.

\begin{figure}[ht]
    \centering
    \includegraphics[width=\linewidth]{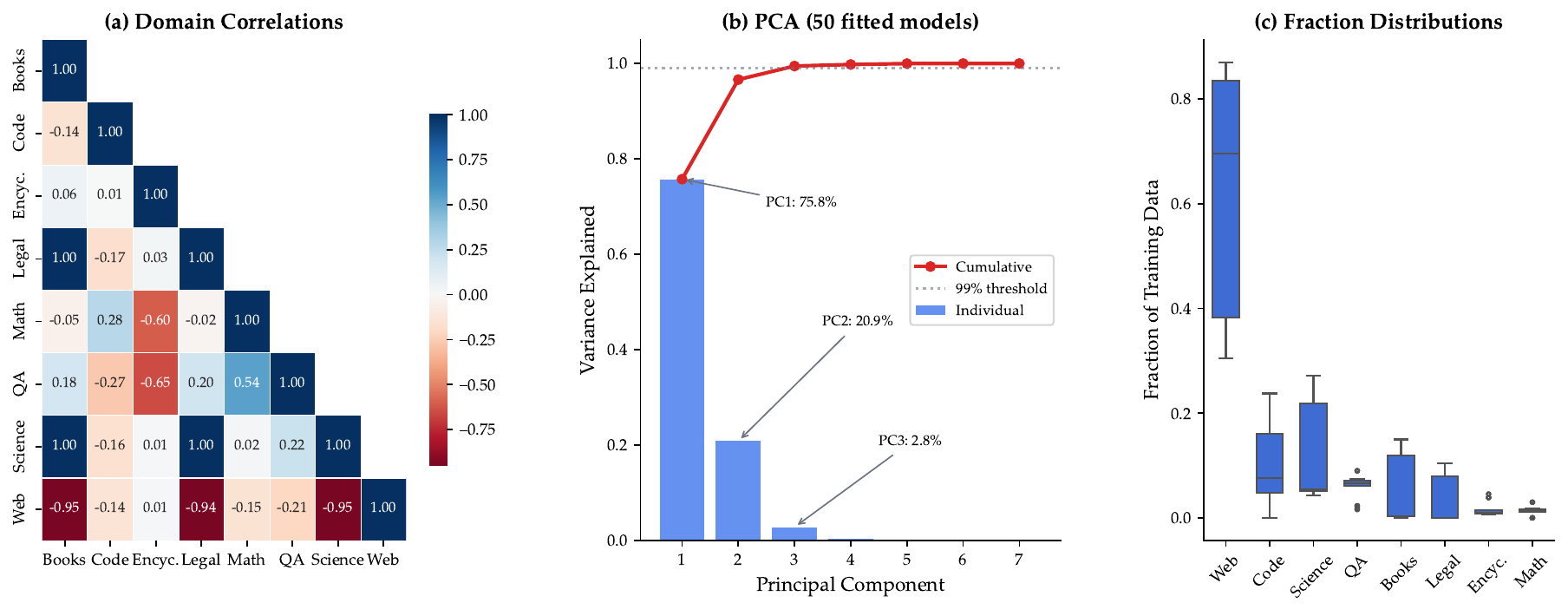}
    \vspace{-0.25in}
    \caption{\textbf{Structure of the composition space.} \emph{\textbf{(a)}} Pairwise correlations between domain fractions: \emph{books}, \emph{legal}, and \emph{science} are correlated. \emph{\textbf{(b)}} PCA scree plot: two components explain 95\% of the variance. \emph{\textbf{(c)}} Distribution of domain fractions: \emph{web} dominates, while \emph{math} and \emph{encyclopedia} never exceed a few percent.}
    \label{fig:domain_structure}
\end{figure}

\paragraph{Compositional structure.}
Although our models span 8 pretraining domains, the observed mixtures span only two effective directions. The first two PCA components explain 95\% of the variance, the first set by code fraction and the second by the correlated books, science, and legal domains. The two sets of models built on the Pile, \texttt{Pythia} and \texttt{GPT-Neo/J/NeoX}, share one diverse mixture (around $30\%$ web, $27\%$ science, $15\%$ books), while the other models are mostly web ($69$ to $97\%$). Our $\gamma$ estimates track the axes along which the models differ (Figure~\ref{fig:domain_structure}).

\paragraph{Identifiability of $\gamma$ vs.\ $(\alpha, \beta)$.}
Across our models, $\log D$ and $\log N$ are correlated ($r{=}0.62$, $R^2{=}0.39$), which creates a ridge in the Chinchilla loss surface: $\beta\log D$ and $\alpha\log N$ trade off freely, and our analysis confirms that within the good-fit region ($\alpha\in[0.4,\,0.9]$), the best-fit $\beta$ varies by an order of magnitude with nearly identical $R^2$. The composition parameters do \emph{not} share this problem. The features $z_{i,k}=u_{i,k}\log(u_{i,k}d_i)$ are nearly orthogonal to $\log N_i$ ($R^2 < 0.09$), so the synergy coefficients $\gamma$ are well-identified even when $\alpha$ and $\beta$ individually are not. Empirically, $\gamma$ estimates are far more stable across seeds and CV folds than either $\alpha$ or $\beta$.

\paragraph{Second-order synergy estimated from a subset.}
The first principal component is mostly code fraction, so we check whether $\Sigma$ reflects only code. We refit the second-order model on the DataDecide checkpoints alone, whose ablations vary composition along axes other than code. The strongest positive synergy with no code term, Q\&A$\times$Web, is recovered with similar magnitude: $\sigma_{\text{QA,Web}} = {+}2.52$ (90\% CI $[{+}1.59, {+}3.79]$) on this subset, versus ${+}2.39$ ($[{+}0.54, {+}4.23]$) on the full model set, and both exclude zero.

\paragraph{Data repetition.}
The data variable in scaling laws refers to fresh tokens, but some models repeat data. Of the 52 models, 19 use source upsampling, 13 of them from the Pile, which repeats small sources such as Wikipedia and books by $1.5$ to $2.5\times$, and none trains beyond about $1.25$ epochs. The other 33, including the 30 DataDecide checkpoints, use unique data. Repetition is set by each training recipe and does not vary with the mixture, so it does not drive the coefficients, and the DataDecide fit above, which has no repetition, recovers the same synergies.

\paragraph{Stability across model size.}
We use one $\gamma_{j,k}$ for all model sizes, which assumes mixture preferences do not change with scale. To test this, we split the model set at 1B parameters (28 large, 24 small) and refit the first-order model on the large half. Of the 13 largest significant $\gamma$, all keep their sign and 9 keep bootstrap 90\% intervals that exclude zero (Table~\ref{tab:size_stability}). Magnitudes shift with the smaller sample, but the ordering of preferred domains holds, so one $\gamma$ per pair is enough over the scales we study. We leave a scale dependent $\gamma(N)$ to future work.

\begin{table}[H]
\centering
\small
\caption{First-order coefficients $\gamma_{j,k}$ on the full model set versus the large half (${\geq}1$B). Signs are preserved even as magnitudes shift.}
\label{tab:size_stability}
\begin{tabular}{lccc}
\toprule
Coefficient $\gamma_{j,k}$ & Full set & Large half & 90\% CI \\
\midrule
code $\to$ \texttt{IFEval}            & $+0.88$ & $+0.78$ & $[+0.33,\,+1.22]$ \\
encyclopedia $\to$ \texttt{HellaSwag} & $-0.76$ & $-0.75$ & $[-0.95,\,-0.54]$ \\
science $\to$ \texttt{ARC-Challenge}  & $-0.54$ & $-0.18$ & $[-0.21,\,-0.14]$ \\
\bottomrule
\end{tabular}
\end{table}

\paragraph{Synergy versus overlap.}
The synergistic domains, such as code, math, and science, are topically related, so the synergies could instead reflect distributional similarity. We check this directly. For each of the eight domains we sample $5{,}000$ documents from the Pile and RedPajama, embed them with \texttt{all-MiniLM-L6-v2} sentence transformer~\citep{wang2020minilm}, and measure the cosine similarity between domain centroids. Overlap does not track synergy: the two most synergistic pairs, (code, science) and (code, math), both have centroid cosine $0.31$, yet several pairs with no synergy are more similar, such as (web, books) at $0.50$ (Table~\ref{tab:overlap}). 

\begin{table}[H]
\centering
\small
\caption{Centroid cosine similarity between domains versus fitted synergy. The two most synergistic pairs, (code, science) and (code, math), are less similar than several pairs with no synergy.}
\label{tab:overlap}
\begin{tabular}{lcc}
\toprule
Pair & Centroid cosine & Synergy in fit \\
\midrule
(code, science)         & 0.31 & strong positive \\
(code, math)            & 0.31 & strong positive \\
(web, books)            & 0.50 & none \\
(web, encyclopedia)     & 0.39 & none \\
(encyclopedia, science) & 0.36 & weak \\
(books, legal)          & 0.31 & none \\
(code, web)             & 0.24 & none \\
\bottomrule
\end{tabular}
\end{table}

\section{Additional Results}

\subsection{Comparison Results}
\label{app:comparison}
\begin{figure}[H]
    \centering
    \includegraphics[width=\linewidth]{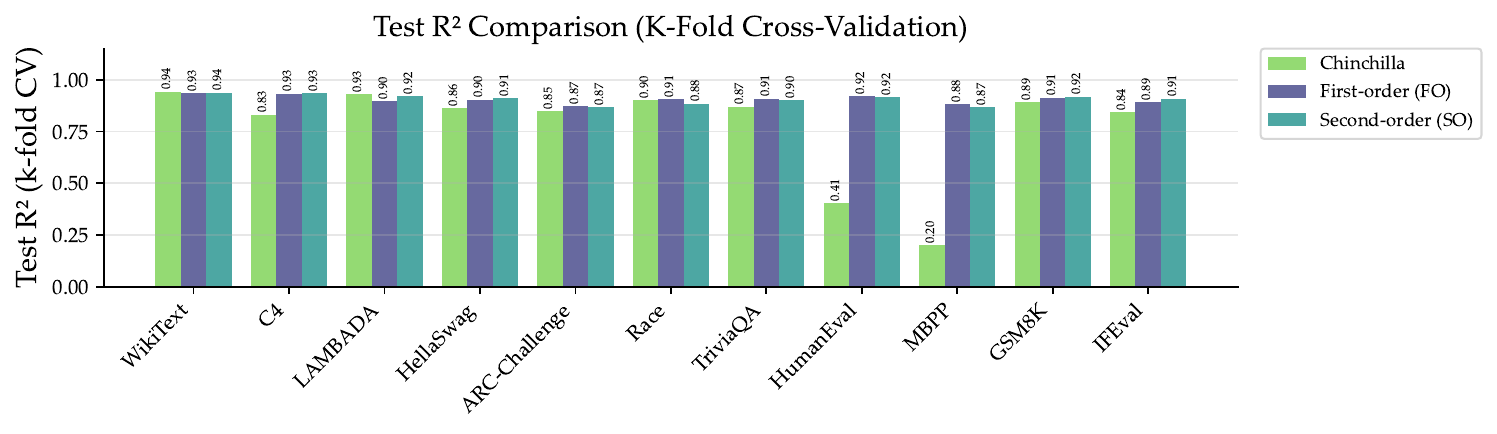}
    \vspace{-0.3in}
    \caption{\textbf{Predictive fit.} Held-out $R^2$ for the Chinchilla baseline (no synergy) and our first-order and second-order models.}
    \label{fig:perf_comparison}

\end{figure}

\begin{figure}[H]
    \centering
    \includegraphics[width=\linewidth]{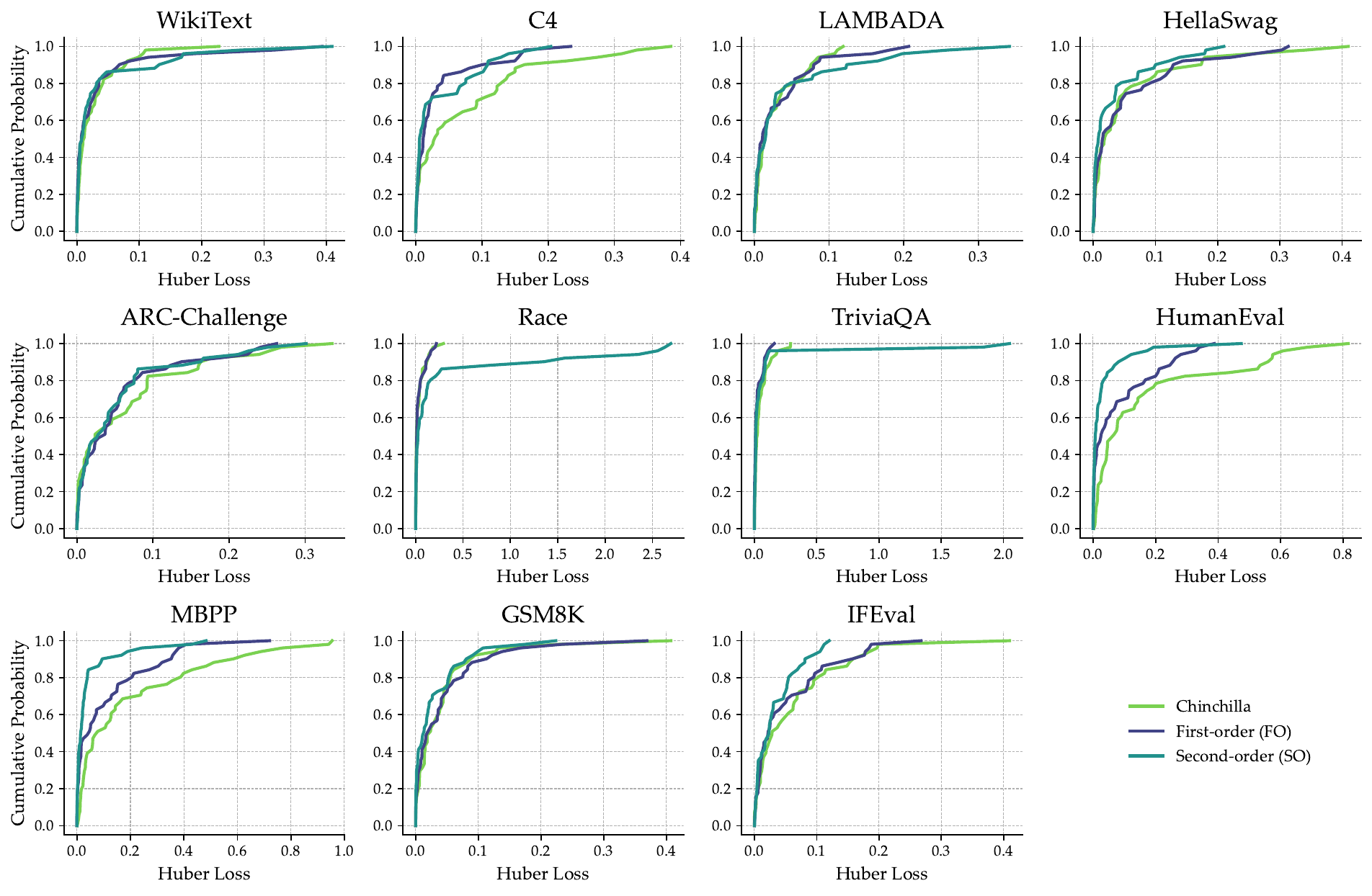}
    \vspace{-0.3in}
    \caption{\textbf{Held-out residual CDF.} Cumulative distribution of held-out residuals for the Chinchilla baseline and our first-order and second-order models. On \texttt{IFEval}, for example, the second-order model has lower held-out Huber loss than the first-order and Chinchilla models.}
    \label{fig:loss_cdf_comparison}
\end{figure}

\begin{figure}[H]
    \centering
    \includegraphics[width=\linewidth]{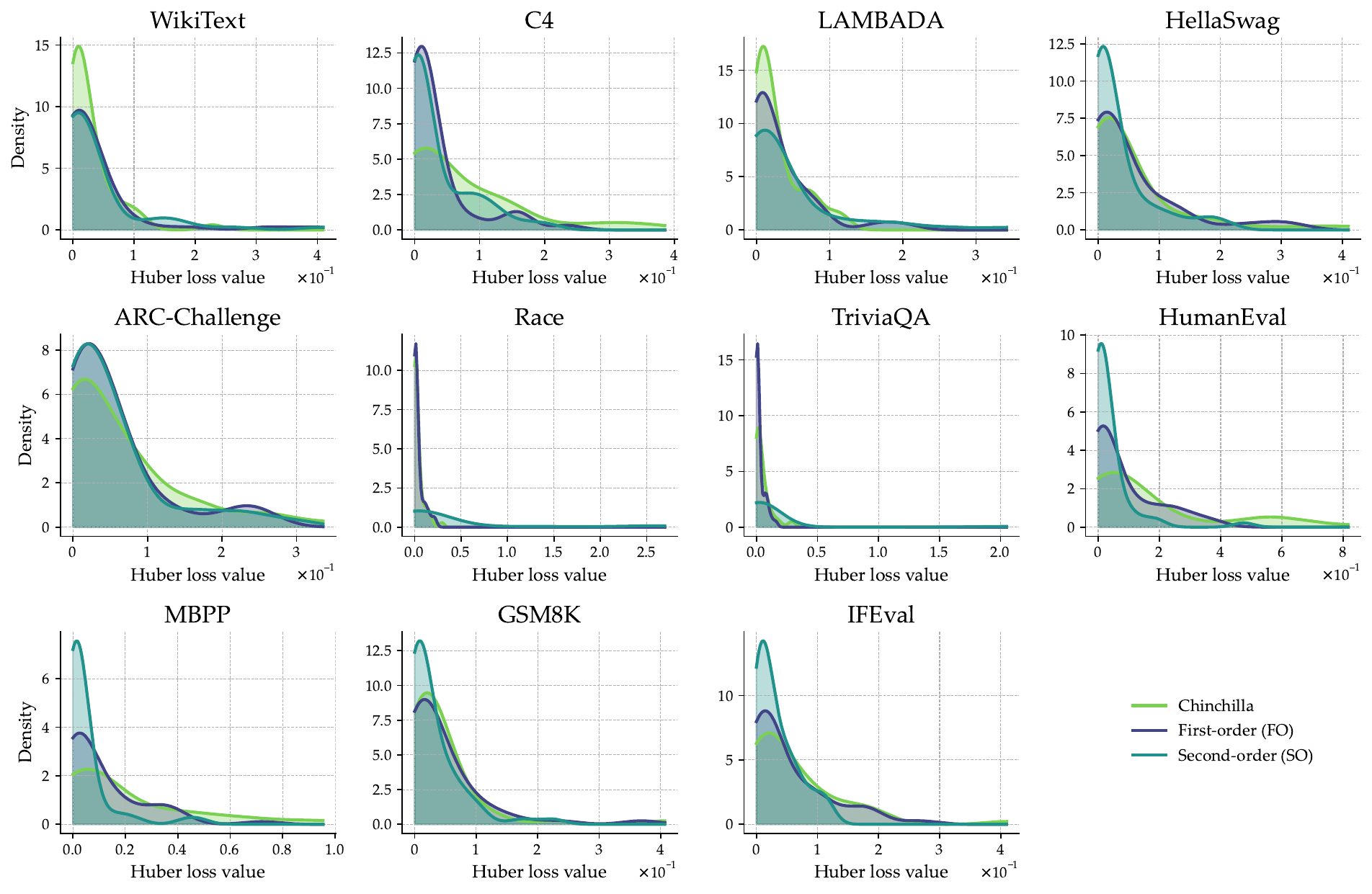}
    \vspace{-0.3in}
    \caption{\textbf{Held-out residual density.} Kernel density of held-out residuals for the Chinchilla baseline and our first-order and second-order models.}
    \label{fig:loss_kde_comparison}
\end{figure}

We assess how predictive our proposed domain-aware scaling law is on held-out samples in more detail. Specifically, we compare the fit of the FO and SO models against the domain-agnostic Chinchilla baseline ($\gamma_{j,k}=0$) on held-out models. Figures~\ref{fig:perf_comparison}, \ref{fig:loss_cdf_comparison}, and \ref{fig:loss_kde_comparison} show $R^2$ and the distribution of held-out residuals for each benchmark. Both synergy models outperform the baseline, and SO gives the best fit on most benchmarks.

\subsection{Scaling Across Models Trained on Different Mixtures}
\label{app:per_family_scaling}

\begin{figure}[H]
    \centering
    \includegraphics[width=\linewidth]{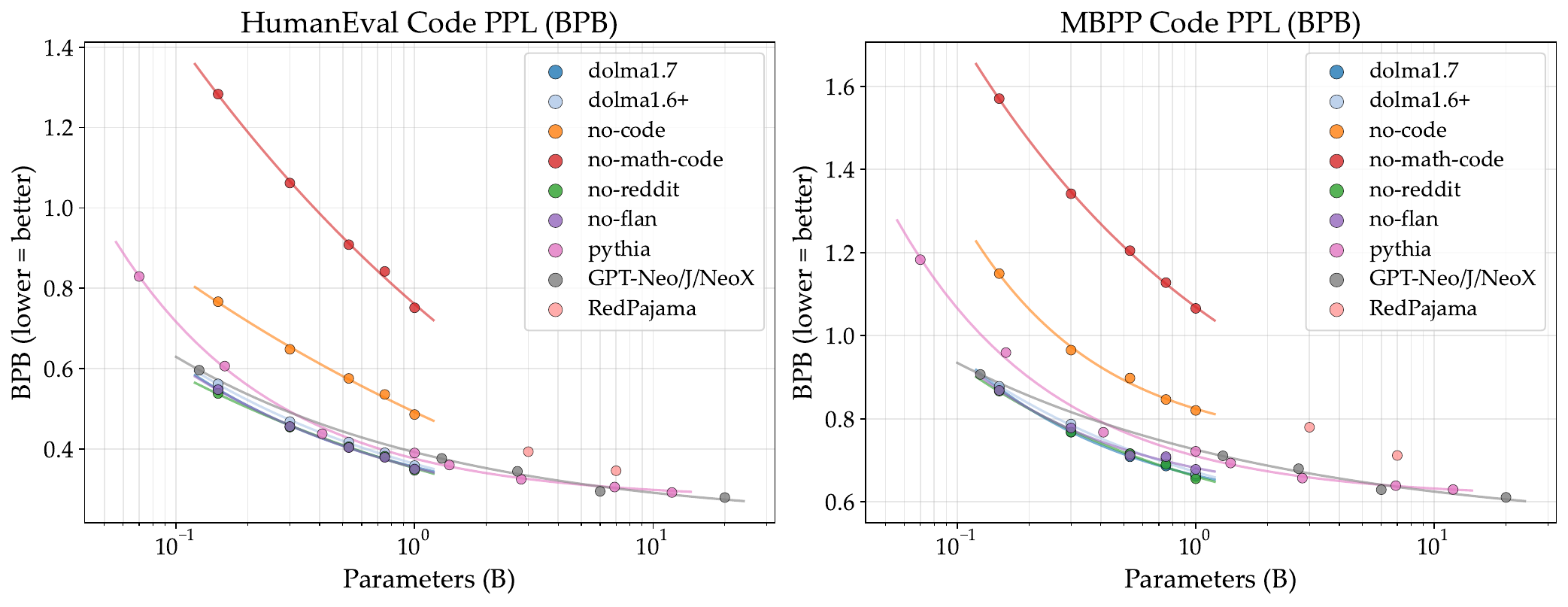}
    \vspace{-0.3in}
    \caption{\textbf{Code scaling by mixture.} Chinchilla fits of code perplexity (BPB) versus model size, one curve per training mixture. \emph{\textbf{Left:}} \texttt{HumanEval}. \emph{\textbf{Right:}} \texttt{MBPP}. The \texttt{DataDecide} ablations (\texttt{dolma1.7}, \texttt{no-code}, \texttt{no-math-code}, \texttt{no-reddit}, \texttt{no-flan}) separate clearly: mixtures with \emph{code} and \emph{math} reach lower BPB at all scales, while removing them shifts the curve up. The external models (\texttt{Pythia}, \texttt{GPT-Neo/J/NeoX}, \texttt{RedPajama}) fall between the \texttt{DataDecide} extremes.}
    \label{fig:per_family_chinchilla_bpb}
\end{figure}

Figure~\ref{fig:per_family_chinchilla_bpb} fits a separate Chinchilla curve to the models trained on each mixture, for code BPB on \texttt{HumanEval} and \texttt{MBPP}. Models trained with \emph{code} and \emph{math} (\texttt{dolma1.7}) sit below those trained without them (\texttt{no-code}, \texttt{no-math-code}) at every size, and the gap holds as models grow. This mirrors the main text result that the \texttt{HumanEval} data exponent drops from $0.45$ to $0.18$ once \emph{code} is removed (Figure~\ref{fig:motivation_fanout}): composition sets the slope of the scaling line, not only its level. \texttt{MBPP} shows the same ordering.

\subsection{Second-order Fitted Coefficients}
\label{app:so_coefficients}

\begin{figure}[H]
    \centering
    \includegraphics[width=\linewidth]{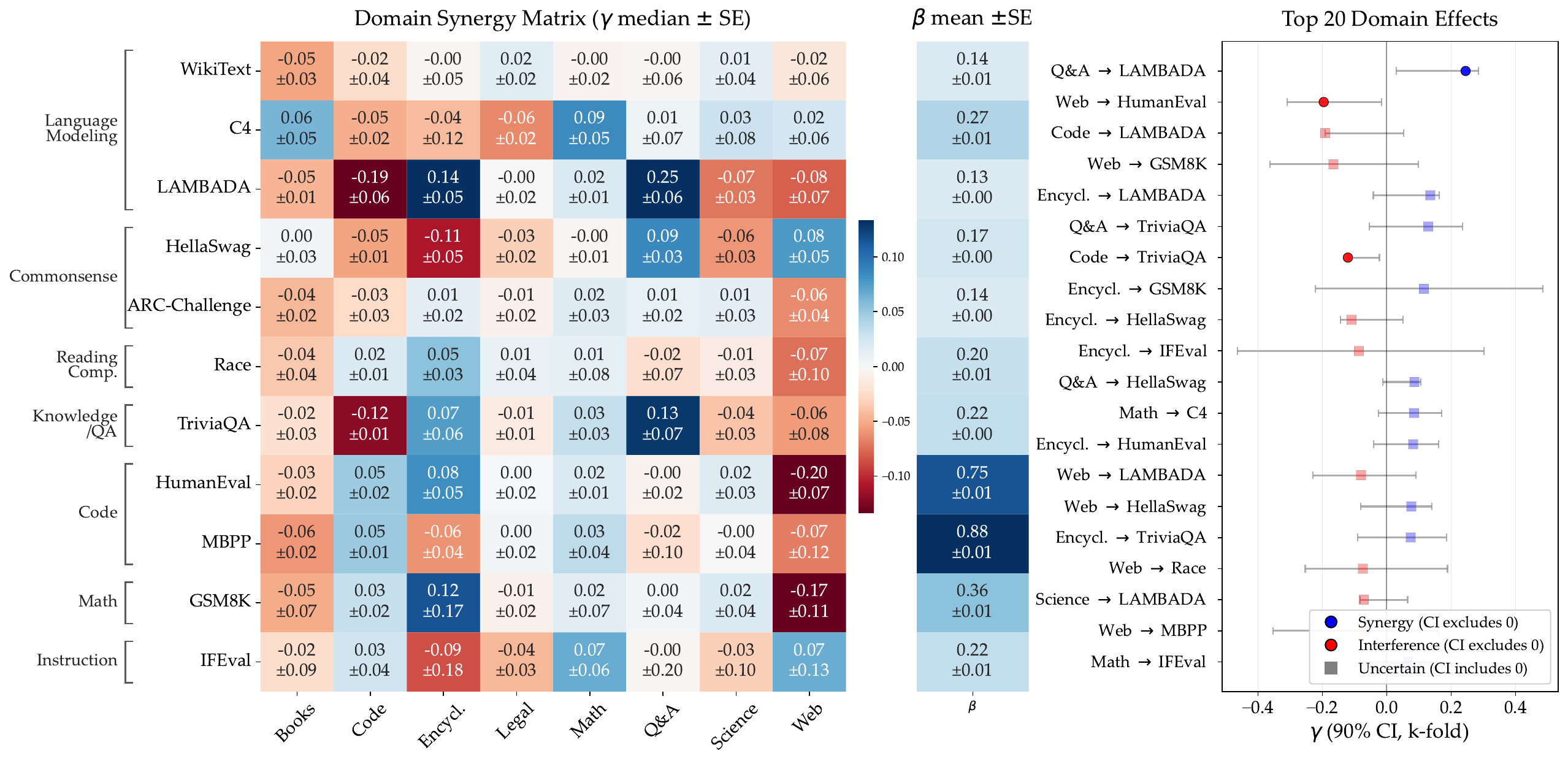}
    \vspace{-0.3in}
    \caption{\textbf{Second-order fitted coefficients.} \emph{\textbf{Left:}} per-domain effects $\gamma_{j,k}$. \emph{\textbf{Center:}} benchmark-specific data exponents $\beta_j$. Both are fit jointly with the shared matrix $\Sigma$ (Figure~\ref{fig:estimated_sigma}); the $\gamma$ shrink relative to the first-order fit because $\Sigma$ absorbs part of the domain interaction.}
    \label{fig:second_order_gamma}
\end{figure}

Figure~\ref{fig:second_order_gamma} reports the per-benchmark coefficients of the SO model, fit jointly with the shared pairwise matrix $\Sigma$ of Figure~\ref{fig:estimated_sigma}.

\section{Comparison with Prior Data Mixture Models}
\label{app:prior_comparison}
Several recent works model how the loss depends on pretraining data composition, but differ in \emph{what} composition effect they model and \emph{how} they are fitted. We first compare the functional forms, then refit each method on our data and compare held-out fit. 

Table~\ref{tab:functional_forms} writes each method in the notation of Section~\ref{sec:problem_setup}, where $u_k$ is the share of pretraining tokens from domain $k$, $N$ is the parameter count, and $D$ is the total token count (we drop the per-model index $i$). RegMix~\citep{liu2024regmix} and Data Mixing Laws~\citep{ye2024data} regress loss on the mixture at a fixed $(N,D)$ and are not grounded in a scaling law over scale. AutoScale~\citep{kang2024autoscale} is an \emph{additive} sum of per-domain power laws, each with its own \emph{fixed} exponent and no parameter count term, with a projection step that extrapolates the optimal mixture across scales. Optimal Data Mixtures~\citep{shukor2025scaling} adds a composition term but keeps the scaling exponents $\alpha,\beta$ fixed across different mixtures.

\begin{table}[h]
\centering
\small
\setlength{\tabcolsep}{6pt}
\renewcommand{\arraystretch}{1.6}
\caption{\textbf{Functional forms of data mixture models}: $u_k$ is the token share of domain $k$, $N$ the parameter count, $D$ the total tokens, $L_\infty$ an irreducible loss, and
subscript $j$ indexes the target benchmark. Remaining coefficients are each method's fitting parameters. RegMix fits $g$ as a nonparametric regressor (gradient-boosted trees) at fixed $(N,D)$. For the second-order row, $s_{kk'}=\operatorname{softmin}_\tau(\bar z_k,\bar z_{k'})$ is the pairwise co-occurrence term of~\eqref{eq:second_order_phi}.}
\label{tab:functional_forms}
\begin{tabular}{ll}
\toprule
Method & Functional form \\
\midrule
Chinchilla~\citep{hoffmann2022training} & $L = L_\infty + A N^{-\alpha} + B D^{-\beta}$ \\
RegMix~\citep{liu2024regmix} & $L = g(u_1,\dots,u_K)$ \\
Data Mixing Laws~\citep{ye2024data} & $L_j = c_j + k_j \exp\!\big(\textstyle\sum_k t_{j,k}\, u_k\big)$ \\
AutoScale~\citep{kang2024autoscale} & $L = L_\infty + \textstyle\sum_k (u_k D + b_k)^{-\mu_k}$ \\
Opt.\ Data Mixtures~\citep{shukor2025scaling} & $L_j = L_\infty + \big(\textstyle\sum_k C_{j,k}\, u_k^{\rho_{j,k}}\big)^{-1} + A N^{-\alpha} + B D^{-\beta}$ \\
\midrule
\textbf{Ours (First-order)} & $L_j = L_{\infty,j} + A_j N^{-\alpha_j} + B_j\, D^{-\beta_j} \prod_k (u_k D)^{-\gamma_{j,k} u_k}$ \\
\textbf{Ours (Second-order)} & FO data term $\times \prod_{k<k'} \exp\!\big[-\beta_j(\gamma_{j,k}{+}\gamma_{j,k'})\,\sigma_{kk'}\, s_{kk'}\big]$ \\
\bottomrule
\end{tabular}
\end{table}

Our framework departs from these in three ways. First, composition enters
\emph{inside the data scaling exponent itself}: the effective per-token return
on benchmark $j$ from domain $k$ is $\beta_j + \gamma_{j,k}$, varying by both
target task and pretraining domain. This is what makes the slopes (not only
the intercepts) of the per-mixture scaling lines diverge in
Figure~\ref{fig:motivation_fanout}. Second, our second-order model introduces an
explicit shared pairwise interaction matrix $\Sigma = \{\sigma_{kk'}\}$ that
captures co-occurrence effects that vanish when either domain is absent, while none
of the previous functional forms include such a term. Third, all of the above are fit
from \emph{controlled} small-scale training runs deliberately swept over
mixture and scale, and we instead fit from \emph{observational} variation across
open-weight LLMs.

\paragraph{Empirical comparison.}
We refit each prior method on our data and compare held-out $R^2$. Every row of Table~\ref{tab:empirical_comparison} uses the same models, benchmarks, and cross-validation as Section~\ref{sec:setup}, and only the predictor changes. Data Mixing Laws and RegMix were designed for a fixed $(N,D)$ and do not include any scale features in their published form, so they collapse to near-zero $R^2$ on our data, where $N$ and $D$ both vary. The \emph{ext.} rows add standardized $\log N$ and $\log D$ to make them comparable. AutoScale is a per-domain power law in the per-domain token count, so its \emph{as published} row already uses the total budget $D$ and reaches $R^2 = 0.43$ (median), though it has no parameter count term. Its \emph{ext.} row adds standardized $\log N$, $\log D$, and an intercept. Even after these modifications, our first- and second-order forms give the best median, mean, and worst-case $R^2$, and the margin is largest in the worst-case column.

\begin{table}[H]
\centering
\small
\caption{\textbf{Held-out $R^2$ of data-mixture models after refitting.} The
\emph{ext.} rows add standardized $\log N$ and $\log D$ to methods that do not include
scale terms.}
\label{tab:empirical_comparison}
\begin{tabular}{lccc}
\toprule
Method & Median $R^2$ & Mean $R^2$ & Min $R^2$ \\
\midrule
\textbf{Second-order (ours)} & \textbf{0.912} & 0.903 & 0.866 \\
\textbf{First-order (ours)}  & 0.906 & \textbf{0.906} & \textbf{0.874} \\
\midrule
Data Mixing Laws (ext.)~\citep{ye2024data}         & 0.910 & 0.871 & 0.554 \\
AutoScale (ext.)~\citep{kang2024autoscale}          & 0.891 & 0.852 & 0.598 \\
Scaling Laws for Optimal Data Mixtures~\citep{shukor2025scaling} & 0.847 & 0.791 & 0.395 \\
RegMix (ext.)~\citep{liu2024regmix}                 & 0.790 & 0.747 & 0.511 \\
\midrule
AutoScale (as published)~\citep{kang2024autoscale}  & 0.430 & 0.426 & 0.232 \\
Data Mixing Laws (as published)~\citep{ye2024data}  & 0.024 & 0.014 & $-0.178$ \\
RegMix (as published)~\citep{liu2024regmix}         & $-0.080$ & $-0.074$ & $-0.185$ \\
\bottomrule
\end{tabular}
\end{table}

\section{Mixture Optimization and Validation}
\label{app:mixture_opt_val}

This appendix expands on the mixture validation experiments of Section~\ref{sec:pretrain}. Appendix~\ref{app:optim_math} derives the objective used to choose the validation mixtures, and Appendix~\ref{app:validation} provides more details for this experiment.

\subsection{Mixture Optimization Derivation}
\label{app:optim_math}

We derive the mixture objective that scores and ranks candidate mixtures for each target task.
Fix a target benchmark $j$, a model size $N$, and a token budget $D$. Write $\mathbf{c}=(c_1,\dots,c_K)$ for a candidate mixture over the $K$ validation domains. For this mixture, the first-order law predicts the log-loss
\[
\operatorname{LSE}\!\bigl(e_j,\; a_j-\alpha_j\log N,\; \Phi_j(\mathbf{c})\bigr),
\qquad
\Phi_j(\mathbf{c})
=b_j-\sum_k\bigl(\beta_j+\gamma_{j,k}\bigr)\,h(c_k)-\beta_j H(\mathbf{c}),
\]
where $h(c_k)=c_k\log(c_kD)$ is the first-order feature $z_k$ from \eqref{eq:domain_benchmark_lse} evaluated at mixture $\mathbf{c}$, and
$H(\mathbf{c})=-\sum_k c_k\log c_k$. 
The first two arguments do not depend on the mixture, and LSE increases in each argument, so the best mixture is the one that makes $\Phi_j$ smallest. Inside $\Phi_j$, the $\beta_j$ terms add up to the same constant for every mixture, because the weights sum to one and \eqref{eq:log_d_decomp} gives $\sum_k h(c_k)+H(\mathbf{c})=\log D$. What remains is the only part the mixture controls:
\[
\Phi_j(\mathbf{c})
= b_j-\beta_j\log D-F_j(\mathbf{c}),
\qquad
F_j(\mathbf{c})=\sum_k\gamma_{j,k}\,h(c_k).
\]
Thus the full single-task mixture optimization is
\[
\mathcal C
=\Bigl\{\mathbf{c}\in\mathbb{R}_{\ge 0}^{K}:
\sum_k c_k=1,\; c_{\min}\le c_k\le c_{\max}\Bigr\},
\]
with objectives
\[
\mathbf{c}^{\mathrm{opt}}_j
\in \arg\max_{\mathbf{c}\in\mathcal C} F_j(\mathbf{c}),
\qquad
\mathbf{c}^{\mathrm{anti}}_j
\in \arg\min_{\mathbf{c}\in\mathcal C} F_j(\mathbf{c}),
\]
respectively. We obtain the validation mixture weights by solving this constrained optimization for each target task. Intuitively, because $h$ increases on the feasible range ($h'(c)=\log(cD)+1>0$ at our token budgets), the maximum places the largest allowed weights on the domains with the largest coefficients and leaves the rest at the minimum weight $c_{\min}$, and the minimum does the reverse.

\subsection{Validation Experiments}
\label{app:validation}

This section provides the full setup for the mixture validation experiments summarized in Section~\ref{sec:pretrain}, including the fitted coefficients behind the predictions, the construction of each mixture, and the training and evaluation recipe.

\paragraph{Fitted coefficients.}
We take the first-order synergy coefficients from our main fit (Figure~\ref{fig:estimated_gamma}) for the three target tasks, \texttt{HumanEval}, \texttt{GSM8K}, and \texttt{IFEval}, restricted to the seven domains (Table~\ref{tab:gamma_predictions}). Math and code are the top positive domains across tasks: math has the largest coefficient for \texttt{HumanEval} and \texttt{GSM8K}, while code leads for \texttt{IFEval}. Books is consistently the most negative across all three tasks.

\begin{table}[H]
\centering
\caption{First-order synergy coefficients $\gamma_{j,k}$ from the fitted scaling law. Positive values indicate that the domain accelerates scaling on the target task. Negative values indicate interference. 5-fold CV $R^2$: \texttt{HumanEval}\,=\,0.921, \texttt{GSM8K}\,=\,0.913, \texttt{IFEval}\,=\,0.892.}
\label{tab:gamma_predictions}
\small
\begin{tabular}{lccc}
\toprule
Domain $\gamma$ & \texttt{HumanEval} & \texttt{GSM8K} & \texttt{IFEval} \\
\midrule
Math         & \textbf{+1.336} & \textbf{+0.668} & +0.307 \\
Code         & +0.467          & +0.186          & \textbf{+0.875} \\
Science      & $-0.082$        & $-0.005$        & $-0.217$ \\
Q\&A         & +0.054          & $-0.018$        & +0.139 \\
Encyclopedia & $-0.689$        & +0.239          & $-0.095$ \\
Web          & $-0.446$        & $-0.234$        & $-0.088$ \\
Books        & $\mathbf{-0.903}$ & $\mathbf{-0.672}$ & $\mathbf{-0.322}$ \\
\bottomrule
\end{tabular}
\end{table}

\paragraph{Mixture construction.}
For the validation runs, we train a simple two-dominant-domain construction: every domain keeps a minimum weight ($c_{\min}=0.02$), one dominant domain receives $c_{\max}$, and a second receives the remaining mass. The cap $c_{\max}$ is set by the token count available for each domain relative to the training budget ($c_{\max}\approx 0.50$ in practice). The resulting predicted-optimal, predicted-anti-optimal, and balanced mixtures are shown in Table~\ref{tab:experiment_mixtures}.

\begin{table}[H]
\centering
\caption{Predicted-optimal, predicted-anti-optimal, and balanced mixtures for each target task. Non-target domains receive 2\% weight.}
\label{tab:experiment_mixtures}
\small
\begin{tabular}{llll}
\toprule
Name & Mixture & Target & Prediction \\
\midrule
humaneval-opt  & math 50\%, code 40\%  & \texttt{HumanEval} & best \\
humaneval-anti & books 50\%, ency 40\% & \texttt{HumanEval} & worst \\
gsm8k-opt      & math 50\%, ency 40\%  & \texttt{GSM8K}     & best \\
gsm8k-anti     & books 50\%, web 40\%  & \texttt{GSM8K}     & worst \\
ifeval-opt     & code 50\%, math 40\%  & \texttt{IFEval}    & best \\
ifeval-anti    & books 50\%, sci 40\%  & \texttt{IFEval}    & worst \\
balanced       & $\nicefrac{1}{7}$ each &                   & baseline \\
\bottomrule
\end{tabular}
\end{table}

\paragraph{Training details.}
We train transformer language models with the OLMo-core~\citep{olmo20242olmo2furious} framework on a single A100 GPU per run, at two scales: 30M-parameter models on 3B tokens and 150M-parameter models on 5B tokens. The 30M model uses $d_\text{model}=256$, 8 heads, and 8 layers, while the 150M model uses $d_\text{model}=640$, 10 heads, and 12 layers. Both use sequence length 2048 and AdamW with Chinchilla scaled learning rates.

Training data comes from the DataDecide project~\citep{magnusson2025datadecide}, pre-tokenized with the \texttt{gpt-neox-olmo-dolma-v1\_5} tokenizer (vocab size 50{,}304). The pretraining data maps mixtures to 7 domains: \textbf{books} (Dolma books), \textbf{code} (StarCoder), \textbf{encyclopedia} (Dolma wiki), \textbf{math} (open-web-math + algebraic-stack from Proof-Pile-2), \textbf{Q\&A} (StackExchange via RedPajama), \textbf{science} (peS2o papers), and \textbf{web} (C4). Since no legal slice is included in this validation data, we do not consider synergies with the legal domain in the mixture optimization (Appendix~\ref{app:obs_data}), and candidate mixtures are scored under the fitted eight-domain law with the legal weight set to zero.

\paragraph{Results.}
Table~\ref{tab:bpb_results} in Section~\ref{sec:pretrain} reports bits-per-byte on each target task's held-out evaluation set at both scales, together with the full discussion. The predicted-optimal mixture reaches lower BPB than the predicted-anti-optimal mixture for all three tasks at both scales, which confirms the directional predictions of the fitted law.